\pdfoutput=1
\documentclass[11pt]{article}

\usepackage{acl}

\usepackage{times}
\usepackage{latexsym}
\usepackage{graphicx}
\usepackage{amsmath}
\usepackage{makecell}
\usepackage{booktabs}
\usepackage{caption}
\usepackage{subcaption}
\usepackage{multirow}
\usepackage[T1]{fontenc}
\usepackage{amsmath}
\usepackage[utf8]{inputenc}

\usepackage{microtype}
\usepackage{inconsolata}

\newcommand{\pledgekwmodel}[1] {PLEDGE-KW-{#1}edits}
\newcommand{\pledgestructmodel}[1] {PLEDGE-Struct-{#1}edits}
\newcommand{\pledgefullmodel}[1] {PLEDGE-Full-{#1}edits}
\title{Investigating Content Planning for Navigating Trade-offs in Knowledge-Grounded Dialogue}
\author{Kushal Chawla$^{1}$\thanks{\hspace{0.8mm} Work done during an internship at Google}\hspace{0.3cm}Hannah Rashkin$^{2}$\\\textbf{Gaurav Singh Tomar}$^{2}$\hspace{0.3cm}\textbf{David Reitter}$^2$ \\
$^{1}$University of Southern California\hspace{0.3cm}$^2$Google DeepMind \\
$^1$\texttt{kchawla@usc.edu}\hspace{0.3cm}$^2$\texttt{\{hrashkin,gtomar,reitter\}@google.com}}

\begin{document}
\maketitle
\begin{abstract}
Knowledge-grounded dialogue generation is a challenging task because it requires satisfying two fundamental, yet often competing constraints: being responsive in a manner that is \textit{specific} to what the conversation partner has said while also being \textit{attributable} to an underlying source document. In this work, we bring this trade-off between these two objectives (\textit{specificity} and \textit{attribution}) to light, and ask the question: Can explicit content planning before the response generation help the model to address this challenge? To answer this question, we design a framework called PLEDGE, which allows us to experiment with various plan variables explored in prior work supporting both metric-agnostic and metric-aware approaches. While content planning shows promise, our results on whether it can actually help to navigate this trade-off are mixed -- planning mechanisms that are metric-aware (use automatic metrics during training) are better at automatic evaluations but underperform in human judgment compared to metric-agnostic mechanisms. We discuss how this may be caused by over-fitting to automatic metrics, and the need for future work to better calibrate these metrics towards human judgment. We hope the observations from our analysis will inform future work that aims to apply content planning in this context.

% Content planning turns out to be a useful tool in both human judgment and automatic metrics. However, planning mechanisms that are metric-aware (use automatic metrics in their training time) are better at automatic evaluations but underperform in human judgment compared to metric-agnostic planning mechanisms. We discuss how this may be caused by over-fitting to automatic metrics, and the need for future work to better calibrate these metrics towards human judgment.
% Reducing hallucinations, or being faithful to the input evidence, is a fundamental challenge for making progress in knowledge-grounded dialogue. Prior work suggests that the problem arises due to both the noise in the collected data, and the developed models that amplify it. Existing approaches that tackle faithfulness tend to converge towards extractive solutions that often use the input knowledge out of context. To this end, we design PLEDGE, a novel controllable and explainable framework for constrained text generation. PLEDGE performs planning for response-generation based on structural and lexical attributes, using interpretable plan-level edits to better satisfy the desirable quality constraints. We perform a comprehensive evaluation on the Wizard of Wikipedia dataset and show that PLEDGE better handles the faithfulness-coherency trade-off compared to baselines, by generating both informative and contextually appropriate responses grounded in the given input evidence.
\end{abstract}

\section{Introduction}
%In recent years, there has been a flurry of attention for knowledge-grounded dialogue tasks with papers describing new datasets \cite{dinan2018wizard,qrecc}, models \cite{shuster2020dialogue,rashkin-etal-2021-increasing,?}, and evaluation methods \cite{rashkin-etal-2023-ais,honovich-etal-2021-q2}.
A knowledge-grounded dialogue system that aims to address a user's information needs must meet two fundamental requirements. First, the knowledge shared by the system must be credible. A common formulation for this constraint is that the system must share information that is faithful or attributable to the retrieved document (what we refer to as {\it attribution}). More importantly, we argue that for the information to be useful to the user, this credibility (as captured by \textit{attribution}) is insufficient -- the generated response must also make sense in the context of the conversation. It must be \textit{specific}, in the sense that it must fit within the flow of the dialogue (what we refer to as {\it specificity}). This fundamental requirement is what differentiates research in this space from single-turn interactions of a user with a typical search engine.

\begin{figure}
    \centering
    \includegraphics[trim={0in .9in 0in 0in},clip,width=.95\columnwidth]{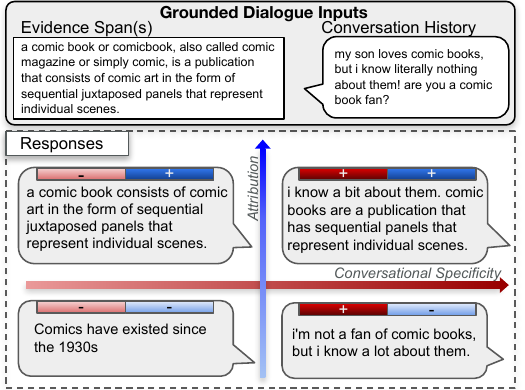}
    \caption{Knowledge-grounded responses need to optimize multiple qualities such as attribution to the evidence document or conversational specificity.} %We investigate planning as an approach for addressing this.
    \label{fig:openingfigure}
\end{figure}

One major open challenge in knowledge-grounded dialogue research is that the model must balance these two objectives, which unfortunately, as we discuss later, can be at odds with each other. For instance, we show in Figure~\ref{fig:openingfigure} how responses can fail along either of these dimensions independently of each other.

% Specificity: 

%  One fundamental requirement to satisfy the user’s information needs is that the knowledge shared by the system must be credible, i.e. the system must share information that is faithful or attributable to the retrieved document (what we refer to as {\it attribution}). 

% , but also, in order to function as a dialogue system, the output needs to reflect what has been said in previous turns without blindly ignoring the conversational context (what we refer to as {\it specificity}).  

There is a scarcity of research explicitly investigating 
%Work in this area may focus on designing models to prioritize optimizing either the attribution or the specificity.  But there isn't much research on
how to navigate the trade-off between these objectives. For example, \citet{rashkin-etal-2021-increasing} investigated using control tokens for improving attribution, but their results showed that this often came at the expense of the specificity of the response to the conversation.  %We demonstrate the issues with optimizing towards only one quality dimension in Figure~\ref{fig:openingfigure}.
In this work, we present a discussion of the challenges in optimizing for {\it both} specificity and attribution in knowledge-grounded dialogue.  In Section~\ref{sec:metrics}, we discuss automatic metrics that can serve as a proxy for these dimensions, demonstrating trivial means to increase either quality at the expense of the other.

\begin{figure*}
    \centering
    \includegraphics[trim={0in 0.35in 0in 0in},clip,width=.9\textwidth]{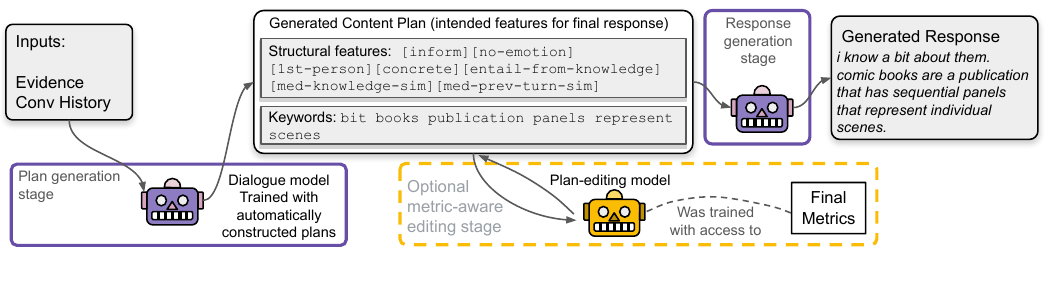}
    \caption{An intuitive overview of the methodology followed in this work to investigate content planning in knowledge-grounded dialogue. We explore plans that use structural variables and keywords.}
    \label{fig:introplanexample}
\end{figure*}

Drawing from other NLG tasks, we pose the following question: \textit{Can explicit content planning help to address this trade-off?} Content planning approaches add an intermediate step of generating the desirable features in the response, referred to as a \textit{plan}, before actually generating the final surface realization conditioned on this \textit{plan}. Prior work showed that splitting the generation into guided steps could be effective in indirectly encouraging the model to be more grounded to commonsense \cite{zhou-etal-2022-think} and source documents \cite{narayan2021planning,narayan2022conditional,hua-wang-2019-sentence}, or to be more coherent \cite{yao-etal-2019-pnw,hu-etal-2022-planet,wu2021semantic,tan-etal-2021}. Hence, it is only natural to hypothesize that content planning can also help to handle the trade-off between these two objectives as well.

To enable a thorough investigation based on various plan variables explored in prior work, we design a framework called PLEDGE. Figure~\ref{fig:introplanexample} provides an intuitive overview of the general methodology followed in PLEDGE. This framework allows us to explore the utility of planning in navigating this trade-off, as well as the effects of structural vs keyword-based plans for this task. While content planning shows promise in general, our results on whether it can actually help to navigate this trade-off are mixed. We observe that planning mechanisms that use automatic metrics during training are better at automatic evaluations but underperform in human judgments compared to mechanisms that do not rely on these metrics explicitly. We discuss how metrics that are better calibrated towards human judgment might help to address this misalignment. We provide insights from our analysis with the hope of informing future work that aims to apply content planning in this context.

We now summarize our contributions: \textbf{I.} We present a computational discussion of the trade-offs between specificity and attribution in knowledge-grounded dialogue (Section~\ref{sec:metrics}), \textbf{II.} We present a novel framework PLEDGE (Section~\ref{sec:methods}) that automates some of the heuristic approaches in prior work to analyze whether content planning can help to handle this trade-off, and \textbf{III.} We present our analysis based on both automated metrics and human evaluation and discuss our insights about the utility of content planning in this context.

% results and takeaways (Section~\ref{sec:expts}) with human judgements and automatic metrics. %showing that while metric-aware planning is useful for automatic evaluations it still lags behind metric-agnostic planning in human judgement. %We close by discussing some of the limitations of metric-aware planning and how it can be improved in future work.

\begin{figure*}
    \centering
\centering
\begin{subfigure}{.5\textwidth}
  \centering
  \includegraphics[width=.98\linewidth]{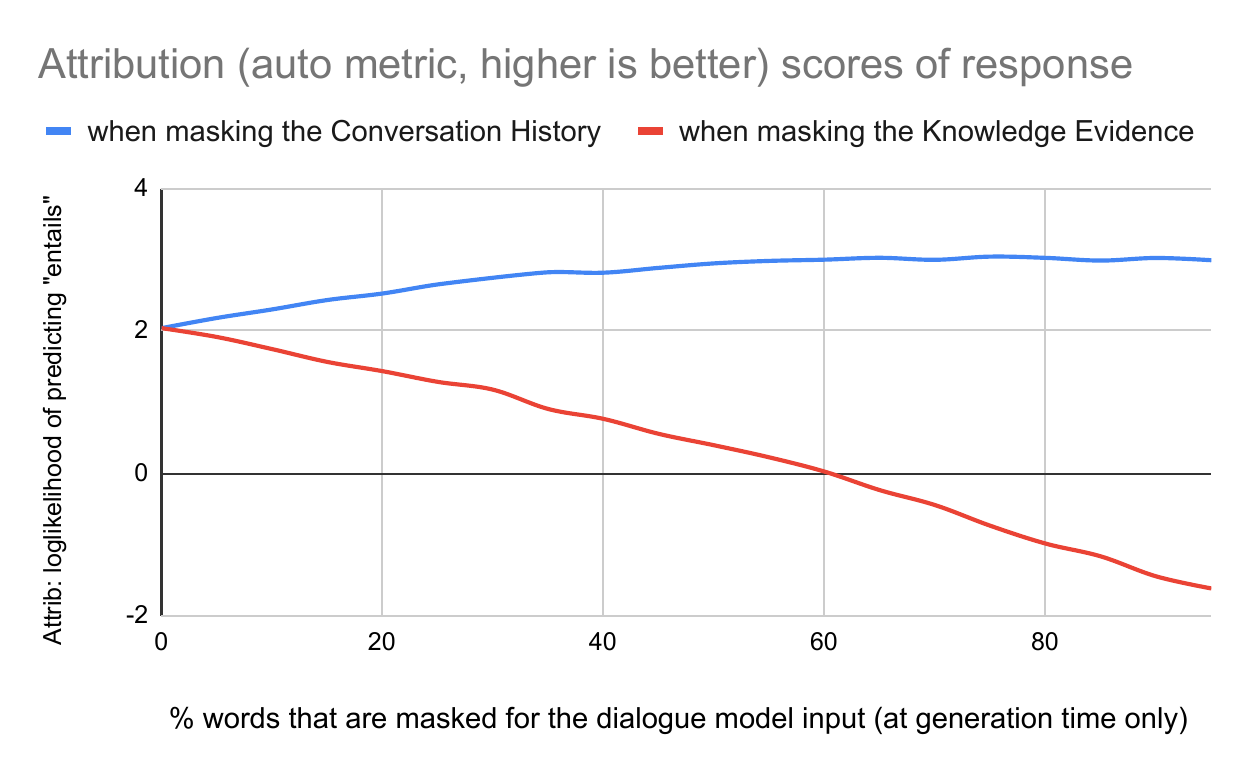}
\end{subfigure}%
\begin{subfigure}{.5\textwidth}
  \centering
  \includegraphics[width=.98\linewidth]{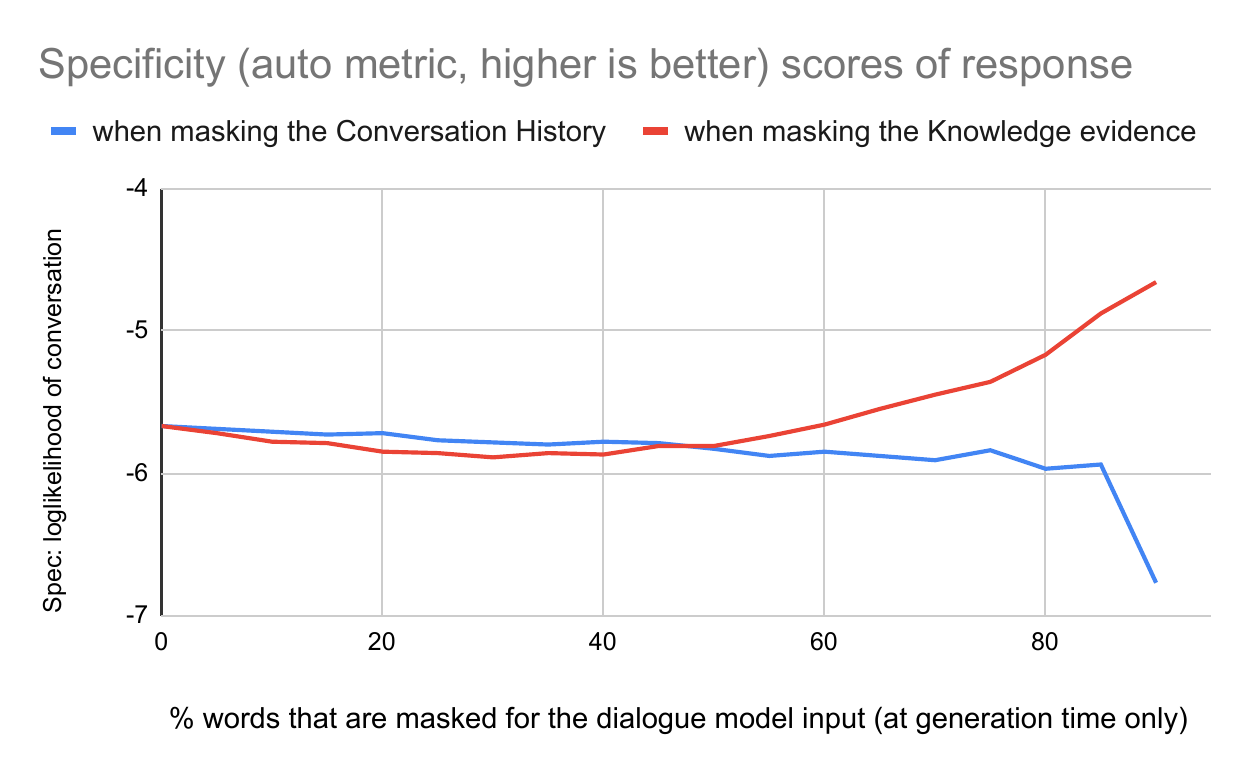}
\end{subfigure}
    \caption{Tradeoff between attribution and specificity scores: We experiment with masking over different portions of the input given to T5.  By simply dropping portions of the evidence or the conversation history, the generated response increases along the specificity or attribution axes respectively, but at the expense of the other score. This shows that these metrics can be gamed when looking at either one in isolation from the other.}
    \label{fig:masking}
\end{figure*}

% \hannah{
% Introduction goes here - what if just follow the story in overview doc? First, trade-offs and issues in knowledge-grounded dialogue to show that we need a framework, then we explore a framework for this problem planning - discuss limitation of prior work in planning, reliance on heuristics for editing, and then advantages of your approach.

% Contributions:
% Faithfulness-coherency trade-off (baselines, dummy models, scatter plot), etc..
% Metrics for coherency - knowledge-grounded settings -> just dgpt without any training seems to make more sense.
% Designing a novel explainable and controllable framework that can tackle the trade-off.
% Planning in knowledge-grounded dialogue -devise plans, expt. with struct and lexical configurations.
% A novel use of text editing in dialogue -> earlier only restricted to the cases with high input-output overlap.
% Showing that our framework can generate informative and contextually appropriate responses for a knowledge-grounded dialogue data WoW.
% Scatter plot or table with slopes magnitudes that shows on automated metrics + human eval 
% Qualitative examples.....
% Changing the plan itself (struct vs lexical vs full), changing the metric for optimization (just coherency, just the factuality, etc), changing the number of edits:- how that impacts the performance, intrinsic metrics for the plans}
\section{Evaluation metrics for grounded dialogue response generation}
\label{sec:metrics}
In the task of knowledge-grounded dialogue, a system  $M_Q$ is given a sequence of previous conversation turns ($x=x_1...x_{n_x}$) and an evidence span  ($e=e_1...e_{n_e}$) selected from a knowledge corpus\footnote{We make the simplifying assumption that an appropriate evidence span has already been labeled.}, and must generate a response $\hat{y}=M_Q(x,e)$ such that the response quality $Q(\hat{y}, x, e)$ is maximized. A good response must be: (1) conversationally appropriate in the context of the rest of the dialogue and (2) accurately representing the information from the knowledge evidence. As mentioned earlier, these two are fundamental to any practically useful knowledge-grounded dialogue system. Hence, we now discuss automated metrics to capture these requirements.

%However, these two objectives can be at odds with each other.  In isolation, either of these objectives can be optimized towards by ignoring the other half of the input (e.g. optimizing faithfulness by just outputting the evidence verbatim while ignoring the conversational history). In this work, we provide a preliminary investigation into a trade-off between these qualities.

\subsection{Metrics approximating attribution to the evidence}
Prior efforts in knowledge-grounded dialogue modeling have often focused on evaluating the faithfulness of responses to evidence \cite{honovich-etal-2021-q2, rashkin-etal-2021-increasing,dziri2022origin}.  In keeping with definitions from related work \cite{rashkin-etal-2023-ais}, we refer to this as {\it attribution} -- a measure of how attributable the information in the response is to the evidence $e$. Such a response conveys knowledge from evidence without hallucinations (information that is not directly inferrable from the provided evidence). This is often estimated by entailment scores from a trained Natural Language Inference (NLI) model.  In this paper, we estimate this with the log-likelihood of predicting entailment using Roberta \cite{roberta} finetuned on MNLI \cite{mnli}).
However, when looked at in isolation from other metrics, maximizing the NLI score is in fact, trivial -- one can simply output the entire evidence span as the response to maximize the entailment scores.  %The difficulty in generating an attributable response is really found in the trade-off with trying to make an attributable response that is also specific to the conversation history.
%Unfortunately, prior models optimized for faithfulness also tend to converge towards this trivial behavior of copying the input evidence (or a part of it) without adhering to the dialogue context (e.g. output generated by control-codes model in Figure~\ref{fig:openingfigure}). To address this concern, in this work, we investigate if incorporating an explicit metric for coherency into the generation process can result in a more desirable behavior.

\subsection{Metrics approximating specificity}
A fundamental requirement for a dialogue system is that the generated response $r$ needs to be conversationally relevant to the previous conversation turns. This is more than topical relevance; the response must follow appropriate conversational discourse and flow logically from the previous turns.  For example, if the previous turn asked a question, it would be inappropriate for the response to not at least acknowledge the question, even if it didn't know the answer.  There are many terms used to describe this dimension of quality  -- {\it relevance}, {\it conversational coherence}, {\it consistency}, and {\it contextual specificity} have all been used in various works to describe related qualities. In this paper, we use the term {\it specificity}, in order to be consistent with a similar dimension set forth by the LaMDA work \cite{lamda}, but we note that this refers to how specific the response is {\it to the conversational history} (not how concrete the language is or other meanings of the word ``specific'').   %Following work by \cite{yeh2021comprehensive}, we initially experimented with various automatic metrics proposed recently in an out-of-the-box manner for coherency on the Wizard-of-Wikipedia dataset to estimate this quality but found that most trained metrics are not suitable for our case because these metrics primarily incentivize lexical overlap in the response and dialogue history, which captures topical relevance more so than discourse between turns. We hypothesize that this is a result of how these metrics are typically trained using heuristically-designed negative samples which are topically misaligned. Hence, 
For our investigation, we use the log-probabilities of response as the next conversation turn using an external dialogue model (the out-of-the-box DialoGPT model \citep{zhang2019dialogpt}) as the most suitable metric to measure coherency. This is similar to how coherence was measured for long text generation in \citet{tan-etal-2021}, which used next sentence prediction probabilities from BERT as a proxy.

\subsection{The trade-off between attribution and specificity}

Because attribution depends on how well the output represents the evidence and specificity depends on how well the output flows from the previous conversation history, we hypothesize that we can increase either of these metrics trivially by forcing a model to attend more to either the evidence or the conversation history.  To test this quantitatively, we use T5-base fine-tuned on Wizard of Wikipedia \cite{dinan2018wizard} data and test on the validation set.  At test time, we apply different levels of dropout on the input words in either the evidence or the conversation history.  As expected, we see in Figure~\ref{fig:masking} that we can increase either the attribution or specificity scores by simply dropping portions of the conversation history or evidence respectively.  However, doing so causes the opposite metric to decrease.  This demonstrates the importance of optimizing for {\it both} when designing new knowledge-grounded response generation models.  Otherwise, when looking at either metric in isolation, it is much easier to game the metric with trivial solutions.  
%%Such a trade-off between being more factual and coherent has also been observed in other scenarios such as text summarization and language modeling \cite{?}. % Hence, an approach for knowledge-grounded dialogue must be capable of jointly optimizing on both of these metrics together, while finding creative ways to address the attribution-specificity trade-off.

For the rest of this work, we judge performance against two extreme cases: one where we trivially maximize the automatic attribution scores by always outputting the evidence verbatim (Attribution-Oracle) and one where we trivially maximize the automatic specificity scores by taking the greedy output of DialogGPT ignoring the evidence (Specificity-Oracle).  In our results section, we normalize the automatic attribution and specificity scores for each model to be scaled between the Attribution-Oracle and Specificity-Oracle scores for easier comparison between the different scales.

\section{Can content planning help?}
\label{sec:methods}

% We investigate different approaches for navigating the trade-off between specificity and attribution, starting with existing methods (Dodecadialogue \cite{shuster2020dialogue}, T5 \cite{raffel2020exploring}, CTRL-T5 \cite{rashkin-etal-2021-increasing}). However, none of these methods were explicitly trained in a way that is optimal for these metrics.

In this work, our goal is to explore whether improved content planning can help with the attribution-specificity trade-off. Content planning has been used in other domains like summarization \cite{narayan2021planning} or chit-chat modeling \cite{zou2021thinking} to help optimize the coherence and attribution of text generations by forcing the model to first ``think'' about what qualities the generated response should have (i.e., choosing a plan $p$) before generating a final surface realization. Prior work has demonstrated that a planning step also adds a layer of inspectability and controllability to the final response \cite{narayan2021planning}. %Based on this recent success in other text generation problems, we hypothesize that content planning can be extended to help knowledge-grounded response generation as well.

More specifically, we aim to answer the following key research questions:

\noindent \textbf{RQ 1:} How helpful is planning out-of-the-box, i.e. without being directly aware of the attribution and specificity metrics that are being optimized?

\noindent \textbf{RQ 2:} How do these metric-agnostic approaches compare with metric-aware methods, where the latter allow explicit optimization towards the desirable quality metrics?

\noindent \textbf{RQ 3:} What kind of structural attributes are useful in the planning stages for this task?

\noindent \textbf{RQ 4:} And finally, is content planning helpful to handle the attribution-specificity trade-off?

To go about answering these questions in a principled manner, we devise a framework called PLEDGE (PLan-EDit-GEnerate). PLEDGE provides an explainable and controllable way to test out various kinds of planning variables explored in prior work, and hence, enables the analysis presented in later sections.

% framework for knowledge-grounded dialogue. The PLEDGE framework helps us to answer the following research questions:

%To answer these questions, we introduce a novel dialogue modeling framework, called PLEDGE (Section \ref{sec:methods:pledge}), and investigate its utility in knowledge-grounded dialogues using automated metrics and human evaluation (Section \ref{sec:expts}).

\section{PLEDGE: PLan-EDit-GEnerate}
\label{sec:methods:pledge}

% In our approach, we define $p$ as a list of plan variables that are intuitively related to the faithfulness or coherence. We also experiment with an approach for allowing the model to edit its plans iteratively to steer the response towards its final objectives ({\it metric-aware editing}).

PLEDGE consists of two modules: a response generation model $G$ (Section \ref{sec:pledge:generation-model}) and an editor $E_Q$ (Section \ref{sec:pledge:plan-editor}). $G$ is our underlying sequence-to-sequence model trained to perform plan-based response generation. The editing model $E_Q$ is tasked with modifying the candidate plans generated by $G$, for better alignment with the quality estimator $Q$. %This module essentially automates the plan-level editing process, replacing the heuristic approaches used in prior work (discussed in detail later). 
Keeping the two modules separate provides the flexibility to train them independently with different datasets and training objectives.

\begin{figure*}[th]
\centering
 \includegraphics[trim={0 1in 0 0 },clip,width=0.95\linewidth]{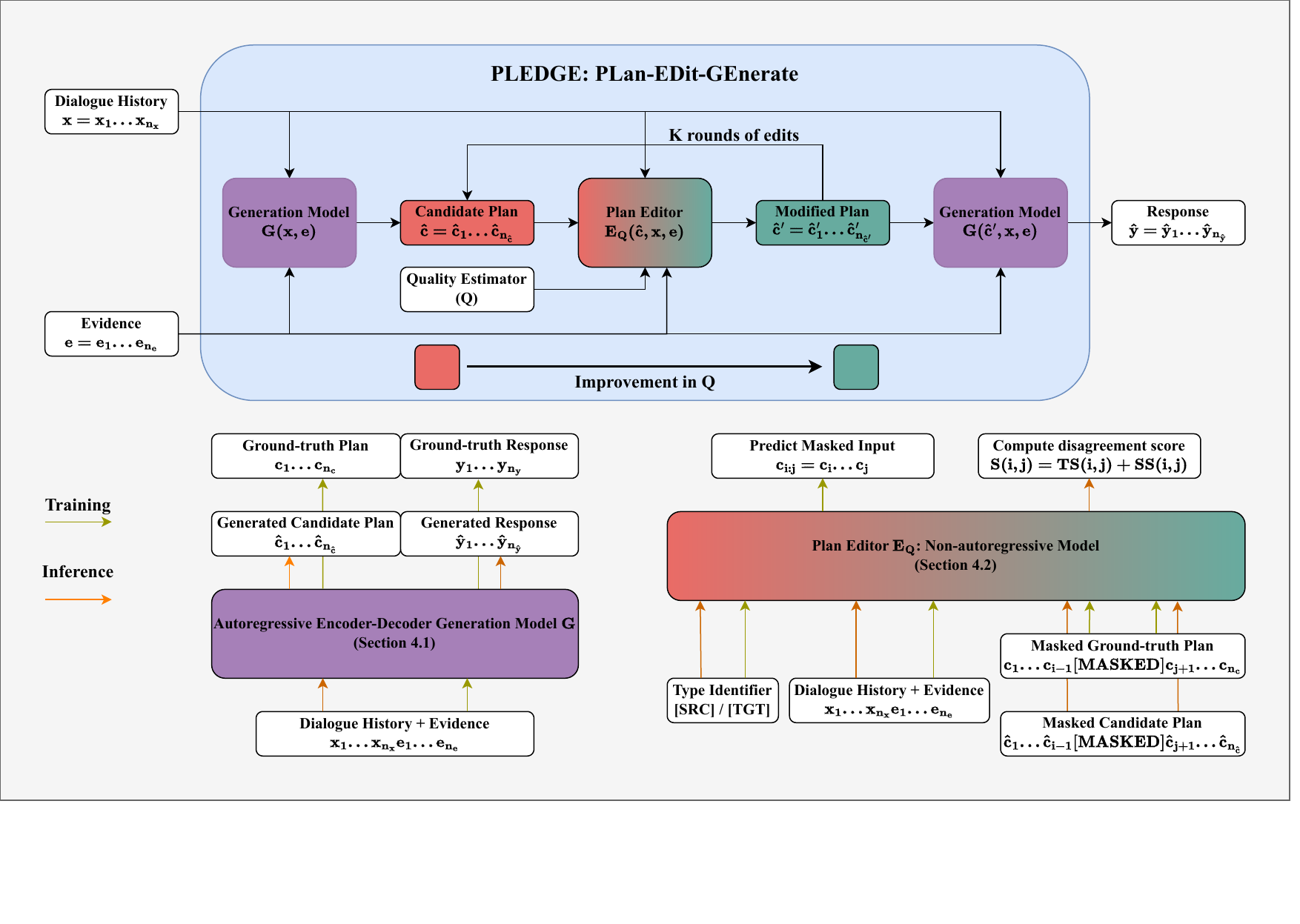}
\caption{Plan-Edit-Generate framework (PLEDGE) -- A general purpose methodology to analyze the benefits of diverse forms of content planning in knowledge-grounded dialogue. PLEDGE consists of two modules -- the primary plan-based response generation model $G$ (Section \ref{sec:pledge:generation-model}, and a plan editing model $E_Q$ that learns to modify a given candidate plan so as to better satisfy the quality estimator $Q$. More details in Section \ref{sec:methods:pledge} and Appendix \ref{sec:plan-editor-appendix}.}
\label{fig:approach_overview}
\end{figure*}

\noindent\textbf{Three-stage inference}: Once $G$ and $E_Q$ are trained, the final response is generated in three stages during inference (top diagram in Figure~\ref{fig:approach_overview}). First, the generation model $G$ takes in the conversation history $x$ and the evidence $e$ to generate a candidate plan $\hat{c} = G(x,e)$. Next, the editor $E_Q$ iteratively modifies this plan to better satisfy the quality constraints defined by $Q$, generating $\hat{c}' = E_Q(\hat{c}, x, e)$. Finally, $\hat{c}'$ is fed back to $G$ to generate the output response $\hat{y} = G(\hat{c}', x, e)$.
%Figure \ref{fig:approach_overview} gives an overview of this process. %We now describe the design of the two modules in detail.

We first describe the general plan format used by our models and then describe the design of the two modules.

\noindent\textbf{Plan Format}: In order to investigate \textbf{RQ 3}, we investigate two different types of plan formats for defining content plans $\hat{c}$. We take inspiration from prior work that used content plans constructed from different kinds of attributes, including dialogue acts, emotion labels, and topic words~\cite{wu2021semantic}, along with phrase outlines~\cite{rashkin2020plotmachines,yao-etal-2019-pnw,tan-etal-2021}, and entity chains~\cite{narayan2021planning}. %For this work, we explore the use of structural attributes (struct), keywords (kw) and both (full) in constructing new plans designed for knowledge-grounded dialogue.
First, we investigate using structural features -- we use a set of variables that describe desired response qualities, such as the level of objectiveness, the proximity to the prior utterance, the proximity to the evidence, dialogue act, and conveyed emotion.  We provide a complete list of these variables along with how they were computed in Appendix \ref{sec:appendix:struct}. We encode each variable using special tokens that we add to the model vocabulary. Second, we investigate a keyword-based plan consisting of an ordered list of the salient words that should appear in the model output (the salient words are selected via tf-idf counts following the keyword-based plan construction procedure proposed by \citet{tan-etal-2021}).
In our experiments, a plan consists of concatenated structural features (struct), a keyword list (kw), or both concatenated with a delimiter (full).  At training time, the plan is extracted automatically from the gold response, and at inference time, they are generated by the generation model.  We include a shortened plan example in Figure~\ref{fig:introplanexample} with more detailed examples in Table~\ref{tab:dataexamples} of Appendix \ref{sec:data-examples-appendix}.

\subsection{Generation Model}
\label{sec:pledge:generation-model}
Our generation model $G$ uses a sequence-to-sequence transformer-based architecture~\cite{vaswani2017attention}, following its subsequent success across a wide range of tasks. We fine-tune the encoder-decoder T5 model~\cite{raffel2020exploring}, although the approach can be trivially extended to a decoder-only design as well. Figure \ref{fig:approach_overview} (bottom left) summarizes how the generation model is designed.

\noindent\textbf{Input}: The input contains the history $x$ and evidence $e$. Both of these sequences are concatenated and fed to the encoder of the seq2seq generation model. See Appendix \ref{sec:io-format-appendix} for more details.

% \footnote{In practice: 1) We separate out the previous utterance from the rest of the dialogue history using a special token. 2) The rest of the conversation turns are added in reverse order with additional special token delimiters to ensure that the most recent information is not lost due to sequence length cutoffs.}

\noindent\textbf{Training}: Before generating the response, the decoder is first trained to generate a \textit{content plan}: a sequence $\hat{c}=\hat{c}_1...\hat{c}_{n_{\hat{c}}}$, conditioned on the encoded input. %The sequence $\hat{c}$ consists of structured attributes that are meant to guide the subsequent response generation. 
After this planning stage, the decoder continues to generate the next ground-truth conversation utterance $\hat{y}=\hat{y}_1...\hat{y}_{n_{\hat{y}}}$, conditioned on the \textit{generated} content plan $\hat{c}$, the \textit{input} conversation history, and the \textit{input} evidence. We train the model for both planning and generation jointly by minimizing the cross-entropy objective for the ground-truth plan sequence $c$ and target utterance $y$:

\begin{equation}
    L_{CE}  = L_{CE}^c + L_{CE}^y,
\end{equation}
where $L_{CE}^c$ and $L_{CE}^y$ are defined as follows:
\begin{equation}
    L_{CE}^c = -\frac{1}{n_c}\sum_{i=1}^{n_c}\textrm{log } p(c_i | c_{<i}, x, e),
\end{equation}
\begin{equation}
    L_{CE}^y = -\frac{1}{n_y}\sum_{i=1}^{n_y}\textrm{log } p(y_i | y_{<i}, c, x, e).
\end{equation}

%%$n_c$ and $n_y$ denote the sequence length for the ground-truth content plan $c$ and target utterance $y$ respectively.

%\noindent \textbf{Data Preprocessing}: Although during inference, the generation model $G$ would be able to itself output a candidate content plan $\hat{c}$ on-the-fly, for training, one needs to have access to the ground-truth sequences $c$. This requires an additional data preprocessing stage which extracts these plans from the ground-truth utterances in the available dialogue data (e.g. by prompting a Large Language Model, statistical measures, trained classifiers, etc.), resulting in quadruples of the form ($x$, $e$, $c$, $y$) for model training.

%The model $G$ provides a straightforward way to answer \textbf{RQ 2}: one can simply vary the construction of the ground-truth plan $c$ before training. Further, 
\noindent\textbf{Inference}: During inference, the same model generates both content plans (conditioned on conversation history and evidence) and the final response (additionally conditioned on the content plan).

The model $G$ by itself is not explicitly optimized towards the desired quality metrics, and hence, provides a metric-agnostic way to incorporate the content plans. Although this will help us answer \textbf{RQ 1}, the model $G$ alone would be insufficient to answer \textbf{RQ 2} which compares metric-agnostic approaches with metric-aware methods.

One way to incorporate the desirable metrics is to apply them in the post-processing stage, once the response is generated by the model $G$. However, these methods often fail to perform the desirable changes in a manner that is still consistent with the input context. Instead, the design of the model $G$ paves the way for another interesting approach to alter the final response - by performing minor alterations to the intermediate plan generated by the model and letting the model itself generate the final response in context. Prior work has relied on heuristics to alter these intermediate plans generated by the model (e.g., by dropping out-of-context keywords). To support our investigation involving diverse planning sequences, we instead need a more generalizable approach. In the next section, we describe an automated way for plan editing -- by tapping into the text editing literature.

\subsection{Plan Editor}
\label{sec:pledge:plan-editor}
We investigate the use of a separate editing model $E_{Q}$, designed to modify a candidate plan sequence to better satisfy the quality estimator $Q$. In practice, this could edit structural variables or add/remove keywords from the plan to push the generation model G to generate a response that would more adequately satisfy some downstream constraint.

We implement our plan editor using the MASKER model \cite{malmi2020unsupervised} from the text editing literature. MASKER provides an unsupervised approach to edit a given input text in a source style $S$ to a target style $T$, by training on \textit{nonparallel} data in the source and target domains ($\theta_{source}$ and $\theta_{target}$). In our case, we are interested in editing plans to enhance the combination of specificity and attribution. Hence, for the source domain data, we select all content plans corresponding to training utterances that score {\it lowly} in the combined automatic attribution and specificity scores (bottom $30\%$ of scores in the training data). The target domain data consists of plans from examples that score {\it highly} in the combined automatic attribution and specificity scores (top $30\%$ of scores in the training data). Otherwise, we use the MASKER model in the same manner as it was originally presented in \citet{malmi2020unsupervised}. We give an overview of the plan editor in Figure \ref{fig:approach_overview}.

\noindent\textbf{Input}: The input consists of a domain identifier ([SRC] or [TGT]), the conversation history $x$, evidence $e$, and a partially-masked plan sequence. During training, this planning sequence comes from the processed ground-truth data, and during inference, this is instead generated by the model $G$.

\noindent\textbf{Training}: The editor relies on a non-autoregressive architecture. While training, the model is fed masked \textit{ground-truth plans} (coming from either the source or the target domain) and is trained to predict the missing plan sequences.

\noindent\textbf{Inference}: During inference, the model simply takes in a \textit{masked candidate plan} and uses the probabilities learned by the model to select an alternative planning sequence that is less probable within the \textit{undesirable} source domain and more probable within the \textit{desirable} target domain (based on what is referred to as the disagreement score).

Since this process follows ~\citet{malmi2020unsupervised}, we only provide a brief overview here. For completeness, we provide more details about the training and inference procedures in Appendix \ref{sec:plan-editor-appendix}.

\section{Experiments}
\label{sec:expts}

We evaluate our models on the Wizard of Wikipedia (WoW) dataset ~\cite{dinan2018wizard} to answer the four \textbf{RQs} from Section \ref{sec:methods}. WoW is a popular dataset consisting of dialogues between an `apprentice', who seeks information, and a `wizard', who has access to relevant documents extracted from Wikipedia. Along with submitting a grounded response, in each turn, the `wizard' also labels the knowledge sentences used for formulating the utterance. We use these labeled sentences as input evidence for the PLEDGE framework. WoW contains $73,571$ instances for training, while $3905$ and $3842$ for validation and testing respectively.\footnote{We mostly report results on the ``seen topic'' portion of the test set since we didn't observe strong differences on the ``seen'' vs ``unseen'' portions.}

\noindent\textbf{Baselines}: We compare to the standard T5 model. We also compare to \citet{rashkin-etal-2021-increasing}, which used T5 with control codes (labeled as ControlCodes in tables) for encouraging attribution but didn't control for specificity.  We also include the baselines (E2E and Dodeca) from that paper.

\noindent\textbf{Training Details}: For all of the models, we use beam-search to be aligned with baselines \cite{dinan2018wizard,shuster2020dialogue}.\footnote{We also experimented with using nucleus sampling \cite{Holtzman2019TheCC} but found that this led to worse attribution scores.} For all variants of planning and controllable models, we used T5-base \cite{raffel2020exploring} as the model architecture for consistency.\footnote{We also tried using T5-large in initial experiments but found similar trends.}  %In keeping with prior work \cite{rashkin-etal-2021-increasing}, we use the oracle retrieval setting, in which we assume that the document retrieval has returned the ``highlighted evidence'' from the Wizard of Wikipedia annotated data at both training and test time.
For training the MASKER model, we used automatically constructed plans from WoW and two different dialogue tasks (TopicalChat \cite{gopalakrishnan19_interspeech} and CMU-DOG \cite{zhou-etal-2018-dataset}). We provide more details in Appendix \ref{sec:expts-training-appendix}. %as one of the benefits of having the separate editor module is that it can be trained on additional data beyond the end-task that the generation model is trained for.

\subsection{Metrics}
As automatic metrics, we report both specificity and attribution as described in the task set-up. As stated in Section~\ref{sec:metrics}, we regularize the scores by scaling linearly between the performance of Attribution-Oracle and Specificity-Oracle. We also report the harmonic mean between these two values as a general measure of the model performance. % ``holistic'' score between the two. %We additionally report BLEU-1,-2,-3,-4 to be in line with prior work.

Additionally, we ran a human evaluation over different model outputs (see Appendix \ref{sec:human-eval-annotation-appendix} for exact phrasing and definitions provided to human annotators) for $100$ examples. 
%The ratings were performed by full-time annotators (3 per example), trained for this specific task, with the supervision of a project manager.  
Annotators ($3$ per example) were first asked to rate the specificity of each model output on a scale of $1$ to $5$ ($5$ being the best), which we scaled between $0$ and $1$ during post-processing. Then, they were asked to rate whether world knowledge conveyed in the response is fully attributable to the evidence (binary question).\footnote{While our work primarily focused on attribution and specificity, we also report human evaluation results on two other metrics (sensibility and interestingness) in Appendix \ref{sec:extra:metrics-appendix}.} % We also asked human to rate a few supplementary qualities (sensibility, interestingness) which we report separately in Section~\ref{sec:extra:metrics}.
In each example, the same annotator viewed the outputs from all of the models first and then annotated each separately.  For the attribution questions, pairs of annotators agreed with each other in $85\%$ of cases. For the specificity questions on the $5$-point Likert scale, pairs of annotator responses on the same output were $\leq$ 1 point from each other in $71\%$ of cases and only strongly disagreed (by $3$ or more points) in 10\% of cases. %The nine models that we included in the annotation study are: the Dodecadialogue model, plain T5, CTRL-T5, metric agnostic/aware\footnote{using 9 edits} Pledge with keywords, structural features, or both.

\subsection{Answering RQ 1 and RQ 2: Metric-Agnostic vs. Metric-Aware Approaches}
\label{sec:expts-rq1-rq2}

\begin{figure}
    \centering
    \includegraphics[width=\columnwidth]{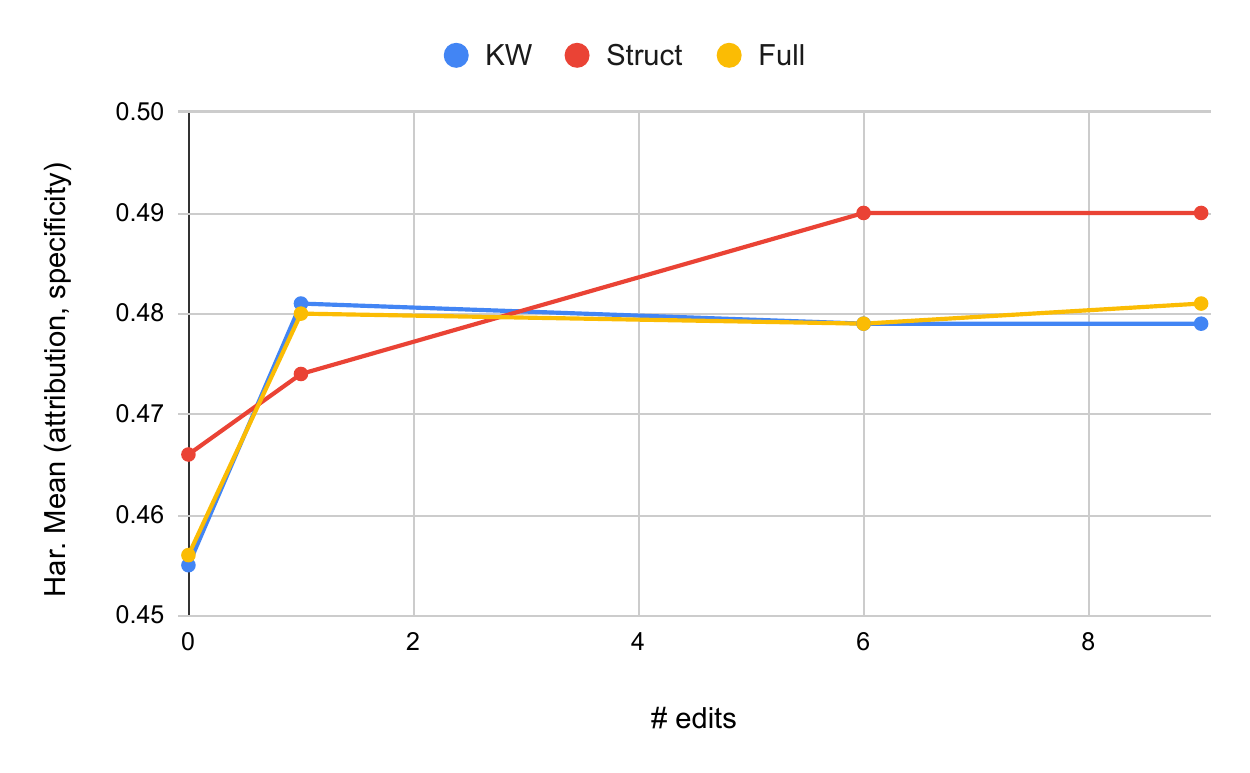}
    \caption{Harmonic mean of attribution and specificity scores increases as plan is edited}
    \label{fig:edits}
\end{figure}
First, we explore the effects of metric-aware editing. We repeat the editing step multiple times and show how the performance changes with the number of edits. We present an editing example in Appendix \ref{sec:plan-editing-examples-appendix}.
Figure~\ref{fig:edits} shows how the harmonic mean of the two automatic metrics improve with the metric-aware editing steps. Generally, the improvements smooth out after about $6$ editing steps.

\begin{table}[t!]
\footnotesize
\centering
\begin{tabular}{l|ccc}
\toprule
& \multicolumn{3}{c}{Human Judgments}\\
\multicolumn{1}{c}{\textbf{Model}} & Specif & Attrib & Hmean \\
\midrule
\pledgekwmodel{0} &	\textcolor{blue}{$\mathbf{0.777}$} &  \textcolor{blue}{$\mathbf{0.873}$}  &  \textcolor{blue}{$\mathbf{0.822}$} \\
\pledgekwmodel{9}&	$0.762$  &  $0.867$  &  $0.811$ \\
\midrule
\pledgestructmodel{0}&	$0.748$  &  $0.830$  & $0.787$ \\
\pledgestructmodel{9}&	\textcolor{red}{$\mathbf{0.719}$}  &  $0.870$ &  $0.787$ \\
\midrule
\pledgefullmodel{0}&	$0.752$  &  $0.837$  &  $0.792$ \\
\pledgefullmodel{9}&	$0.742$  &  \textcolor{red}{$\mathbf{0.813}$}  &  \textcolor{red}{$\mathbf{0.776}$} \\
\bottomrule
\end{tabular}
\caption{\label{tab:humanedits} Human judgements on the seen portions of WoW test set. We report the average attribution and specificity scores (each scaled to be between 0 and 1). We also report the harmonic mean between the two metrics (HMean). The worst and the best scores for each column are in \textcolor{red}{\textbf{red}} and \textcolor{blue}{\textbf{blue}} respectively.}
\end{table}
However, we find different trends in the human evaluations (Table~\ref{tab:humanedits}), where editing rarely improves human judgments. That is, metric-aware edits may be useful for improving the automatic metrics they are trained on, but these improvements do not transfer well to human judgments. This implies that the metric-aware edits may overfit to artifacts in the automatic metrics.  For example, we observe that metric-aware output tends to be shorter and more bland, which may allow it to cheat the specificity metric since the DialogGPT model gives higher likelihood scores to short, bland phrases. For instance, in the example in the appendices, the output generated by the initial plan was ``i'm not sure, but i do know that iguanas can range in length including their tail'', but after editing the new plan leads to the response ``yes they can range in length including their tail'', which is shorter and more generic.  While metric-aware editing would be very useful in situations with better-calibrated automatic metrics, the existing automatic metrics in this space may not be well enough calibrated to act as a proxy for optimizing human judgment.

\subsection{Answering RQ 3: Comparing Different Plan Formats}
\label{sec:expts-rq3}

We generally find that the keyword plan structure is more beneficial than using the structural features in human judgments (Table~\ref{tab:humanedits}).  That said, the structural variables do give the model an advantage in the automatic metrics.  Based on this, we believe that keyword plans may be better for most end-user applications, but structural features may still be useful in specific task setups.

\subsection{Answering RQ 4: Comparison to baselines}
\label{sec:expts-rq4}

\defcitealias{dinan2018wizard}{Di18}
\defcitealias{shuster2020dialogue}{Sh20}
\defcitealias{rashkin-etal-2021-increasing}{Ra21}
\defcitealias{raffel2020exploring}{Ra20}

\begin{table}[t!]
\centering
\scalebox{0.85}{
\begin{tabular}{l|ccc}
\toprule
\multicolumn{1}{c}{\textbf{Model}} & \multicolumn{3}{|c}{Automatic Metrics} \\
& \textbf{Attrib} & \textbf{Spec} & \textbf{HMean} \\ \midrule
Reference & $.189$ & $.297$ & $.231$ \\
Attribution-Oracle & \textcolor{blue}{$\mathbf{1.0}$} & \textcolor{red}{$\mathbf{0.0}$} & \textcolor{red}{$\mathbf{0.0}$}  \\
Specificity-Oracle& \textcolor{red}{$\mathbf{0.0}$} & \textcolor{blue}{$\mathbf{1.0}$} & \textcolor{red}{$\mathbf{0.0}$} \\ \midrule \midrule
\makecell[l]{E2E (\citetalias{dinan2018wizard})}& $.183$ &$.500$ & $.268$  \\
\makecell[l]{Dodeca (\citetalias{shuster2020dialogue})}& $.656$ & $.338$ & $.446$  \\
T5 (\citetalias{raffel2020exploring}) & $.639$ & $.385$ & $.481$ \\
ControlCodes (\citetalias{rashkin-etal-2021-increasing}) & $.862$ & $.297$ & $.442$ \\ \midrule
{\textbf{Plans without Editing}} \\
\pledgekwmodel{0}  & $.595$ & $.368$ & $.455$ \\
\pledgestructmodel{0}& $.543$ & $.409$ & $.466$ \\
\pledgefullmodel{0} & $.520$ & $.406$ & $.456$ \\ \midrule
{\textbf{Plans with Editing}} \\
\pledgekwmodel{9}& $.660$ & $.376$ & $.479$ \\
\pledgestructmodel{9} & $.802$ & $.353$ & \textcolor{blue}{$\mathbf{.490}$} \\
\pledgefullmodel{9}& $.648$ & $.382$ & $.481$ \\ \bottomrule
\end{tabular}}
\caption{\label{tab:autoresults} Results on the seen portions of WoW test set. We report the scaled attribution and specificity scores, and the harmonic mean between the two metrics (HMean). The worst and the best scores for each column are in \textcolor{red}{\textbf{red}} and \textcolor{blue}{\textbf{blue}} respectively. See Appendix \ref{sec:bleu-appendix} for results on BLEU.}
\end{table}

\begin{table}[t!]
\centering
\resizebox{.985\columnwidth}{!}{%
\begin{tabular}{l|ccc}
\toprule
& \multicolumn{3}{c}{Human Judgments}\\
\multicolumn{1}{c|}{\textbf{Model}}  & Spec & Attrib & HMean \\
\midrule
Dodeca & $0.762$ $\pm$ $.017$ &	\textcolor{red}{$\mathbf{0.863}$} $\pm$ $.023$ & $0.809$\\
T5 & $0.761$ $\pm$ $.017$ & $0.880$ $\pm$ $.022$ & $0.816$\\
CTRLCodes & \textcolor{red}{$\mathbf{0.718}$} $\pm$ $.017$ & \textcolor{blue}{$\mathbf{0.907}$}  $\pm$ $.019$ & \textcolor{red}{$\mathbf{0.802}$}\\
PLEDGE-KW & \textcolor{blue}{$\mathbf{0.770}$}  $\pm$ $.016$ &	$0.873$  $\pm$ $.022$ & \textcolor{blue}{$\mathbf{0.822}$}\\
\bottomrule
\end{tabular}%
}
\caption{\label{tab:humanresults} Human judgements on WoW test set. We report average specificity and attribution scores, along with the standard error of the mean (after the $\pm$ symbol). We also report the harmonic mean between the two metrics (HMean). The worst and the best scores for each column are in \textcolor{red}{\textbf{red}} and \textcolor{blue}{\textbf{blue}} respectively.}
\end{table}

\noindent\textbf{Automatic Evaluation}: To get a general insight into whether content planning can help to handle the trade-off, we discuss the strengths and shortcomings of planning in comparison to other methods. The automatic evaluation in Table~\ref{tab:autoresults} shows that most planning models generally outperform most of the baselines on the combined harmonic mean of attribution and specificity. PLEDGE-struct with editing gets the highest combined performance.

Further, the standard seq2seq T5 fine-tuning baseline proves to be a strong model once we combine attribution and specificity. This model outperforms all baseline methods and the plan-based methods that do not use editing on the HMean metric. This observation attests to the idea that only observing one metric can be misleading since both Dodeca and ControlCodes beat the T5 fine-tuning baseline if we only look at the Attribution metric.

Finally, we note that the automated metrics usually rank the system-generated responses higher than human-generated responses (the Reference baseline). We believe that the failure cases for the automatic metrics tend to miss more nuanced linguistic phenomena that appear in more human responses, which often contain richer and more diverse sentence structures. %slightly higher specificity-faithfulness scores on automatic metrics.

\noindent\textbf{Human Evaluation}: We also report results from human evaluations in Table \ref{tab:humanresults}. We only include PLEDGE-KW since it was the highest performer from Table~\ref{tab:humanedits}). The margins between the different models are much smaller than with the automatic metrics, and the trends are slightly different. PLEDGE-KW with keyword-based editing slightly outperforms the other models, albeit not by a significant margin. Further, all models (even with content planning) tend to display a trade-off between specificity and attribution, where the models with higher attribution scores tend to have lower specificity and vice versa. This again underscores that rankings depend on which metric is being prioritized, and future work may need to find more nuanced ways of determining which score is more important on a case-by-case basis.

\noindent\textbf{Qualitative Observations}: We provide sample model outputs in Appendix \ref{sec:example-model-outputs-appendix}. We find that satisfying both criteria of attribution and specificity together can be quite challenging, especially when the input evidence does not directly answer the user query. Often, the models employ ‘yes, and’ type creative improv techniques~\cite{cho-may-2020-yesand}, which acknowledge the previous user utterance while also incorporating the given evidence. Further, the relatively small response length in our dataset hides granular differences between the compared models. Hence, using additional datasets with longer outputs or fine-grained evaluation metrics might be helpful in the future.

\section{Related Work}

\noindent\textbf{Knowledge Grounded Dialogue Evaluation}: Dialogue tasks use multiple dimensions of quality including specificity, sensibleness, and interestingness (SSI) \cite{lamda}, cooperativeness \cite{dziri-etal-2022-faithdial}, in addition to general-purpose text evaluation dimensions regarding clarity, naturalness, and more \cite{howcroft-etal-2020-twenty}.  Knowledge-grounded dialogue tasks require additional metrics of how well the outside evidence is being used in the response, measured as faithfulness or attribution \cite{rashkin-etal-2023-ais}. However, attribution has been noted to have trade-offs with multiple other qualities such as abstractiveness \citep{daheim2023elastic}, engagement \cite{kodama-etal-2023-knowledge}, fluency \cite{aksitov2023characterizing}, and diversity \cite{dziri-etal-2021-neural}. Work in summarization has found similar trade-offs for attribution with abstractiveness \citep{dreyer-etal-2023-evaluating,ladhak-etal-2022-faithful} and diversity \cite{aralikatte-etal-2021-focus,chen-etal-2023-fidelity}. We explore the trade-offs between attribution and specificity for knowledge-grounded dialogue, presenting an analysis of using more explicit planning as a means of mitigation.

\noindent\textbf{Improving different aspects of quality}: Many recent works in knowledge-grounded dialogue tasks \cite{dinan2018wizard,ghazvininejad2018knowledge,gopalakrishnan19_interspeech} have sought to improve attribution scores through many different techniques \citep[e.g. ][\textit{inter alia}]{daheim2023elastic,deng-etal-2023-towards,sun-etal-2023-towards,jang-etal-2023-post,tian2020response}. %, such as elastic weight removal \citep{daheim2023elastic}, focus learning \citep{deng-etal-2023-towards}, contrastive data augmentation \citep{sun-etal-2023-towards}, post-hoc editing \citep{jang-etal-2023-post}, and %using teacher-student frameworks to constructing document memory matrices  \citep{tian2020response}.  
In other open-domain dialogue tasks,  specificity has been improved via better conversation flow like %optimizing for 
smoother transitions \citep{kim-etal-2022-flow,gupta-etal-2022-flow,sevegnani-etal-2021-flow} and the use of ``yes-and'' statements %to drive the conversation forward
\citep{cho-may-2020-yesand}. There are a few recent works that also try to optimize jointly towards both attribution and specificity.  In \citet{nandwani-etal-2023-pointwise}, they use conditional PMI in their decoding strategy for a response that is both faithful to the knowledge spans and relevant to conversational history.  In \citet{wilie2023pick}, they use a rescoring technique that reranks candidate outputs based on the combined score of specificity (relevance score) and attribution (faithfulness score).

\noindent\textbf{Planning for Text Generation}: A plan refers to higher-level reasoning that is used to guide the final text generation, such as for poetry generation \cite{tian2022sonnet}, story generation~\cite{yao-etal-2019-pnw,rashkin2020plotmachines}, text summarization~\cite{narayan2021planning,narayan2022conditional}, or open-domain dialogue~\cite{wu2021semantic,adolphs2021reason,zou2021thinking}. Planning-based neural response generation has shown remarkable promise for adding interpretability to otherwise black-box neural models. Planning improves explainability, by giving insight into the model's decision-making and enhances controllability, by allowing intervention during inference to modify the candidate plans. To the best of our knowledge, our metric-aware editor is the first attempt to handle this intervention automatically, as opposed to relying on heuristics as used in prior work~\cite{narayan2021planning}.

\noindent\textbf{Ongoing Challenges}
In this paper, we use T5-base architectures that are relevant to prior and contemporaneous work \citep{rashkin-etal-2021-increasing,dziri-etal-2022-faithdial,wilie2023pick}.  However, our work is also relevant to more recent large instruction-tuned language models that can struggle with similar issues regarding specificity and attribution. Hallucinations remain an ongoing challenge for LLMs in both conversational QA \cite {chiesurinetal} and other QA settings \cite{adlakhaetal}.  Others have also observed that, while recent modeling progress has led to more fluent and coherent responses overall, these models can still default to stereotypical or generic responses %to dialogue utterances
\cite{zhou-etal-2022-reflect} or fail to be consistent %with prior history
\cite{wangetalforgetting}. Some common mitigation approaches include chain-of-thought reasoning \cite{wangetalcot}, where the model reasons about the response before generation, or post-hoc revision, where errors in the output are corrected after the initial generation \cite{pengetalcritique,baeketal}. %While using different architectures, we see PLEDGE as being conceptually aligned with these approaches.  
PLEDGE has some conceptual similarities to these approaches. Chain-of-thought reasoning is similar to the plan-then-write approach in that the model performs intermediate steps where it reasons about the output before writing. There is also a parallel between revision approaches and the way in which the PLEDGE plan editor module helps correct mistakes from the initial %generative model
output.  While out of the scope of this paper, we hope that future work will investigate how larger instruction-tuned models can leverage our observations about the specificity-attribution trade-off for this task.

%\noindent\textbf{Text Editing}: Our plan editing module relies on the advancements in text editing approaches~\cite{malmi2022text}. These methods convert the task of sequence-to-sequence generation into sequence editing to improve the efficiency, making them suitable for scenarios where the inputs and outputs have high overlap, such as grammatical error correction and style transfer~\cite{mallinson2022edit5}. To employ them for dialogue, we perform editing at the planning stage consisting of higher-level attributes, where a small number of edits lead to high variance in the generated responses. Most text-editing approaches rely on the availability of abundant parallel data for training~\cite{awasthi2019parallel,malmi2019encode,mallinson2020felix,omelianchuk2020gector}. Given the constraints of our problem, we use the MASKER model~\cite{malmi2020unsupervised} which trains a unified model in an unsupervised way without needing any synthetic data and explicit edit operations. Plan-level edits provide an exciting way to extend text-editing to other generation tasks with low input-output overlap.

\section{Conclusion}
We investigated the trade-off between attribution and specificity for knowledge-grounded dialogue, analyzing whether content planning prior to final output generation can help to navigate this trade-off. We find that although content planning shows promise in general, we observe differences in the trends in automated and human evaluations. Hence, whether content planning can help to handle the trade-off remains an open question and more efforts are needed to answer it, with automated metrics that are potentially better calibrated with human judgment. We hope that the insights gained in this work inform future efforts on exploiting content planning in similar contexts.

% requires more work in the future. due to the  We hope that the insights gained in this work guide similar future efforts on exploiting content planning for text generation tasks.

\section{Broader Impact and Ethical Considerations}
We note that we verified the license terms of the datasets used in this work. All the datasets are popular and publicly available for dialogue research.

The primary goal of a knowledge-grounded dialogue system is to be able to converse with a user about the external world, providing the user with important new information. This could lead to dangers of spreading misinformation if a model hallucinates or shares information from untrusted sources. In this work, we put forth attribution metrics as a way of quantifying whether a system hallucinates compared to what was written in the grounding document. However, we make the assumption that the document itself is trustworthy by only using pre-selected document examples from Wikipedia. For more general-purpose systems, more work is needed to quantify the trustworthiness of underlying sources. Additionally, in this paper, we do not evaluate for other important dialogue complications, such as toxic or offensive language, which would need to be taken into account for a real-world dialogue system.

\section{Limitations}
We promote the trade-off between specificity and attribution as an important set of qualities that a dialogue system must ensure, but we acknowledge that this not a sufficient set of qualities that a dialogue system should have. There are other aspects of quality that need further consideration (such as interestingness or different aspects of fluency). Future work may need to extend to exploring complex multi-dimensional trade-offs that go beyond the scope of this work. 

Although we investigate a few different forms of planning mechanisms and how they impact the performance trade-off, there are other forms of planning and guiding structured output that are still largely unexplored for this task. These are beyond the scope of this work, but we encourage future work to explore this direction.

\section*{Acknowledgments}
We would like to thank the anonymous ARR reviewers and the EACL $2024$ Committee for their time in reviewing our work. We thank Chung-ching Chang and Shashi Narayan for their valuable feedback. We are also grateful to Ashwin Kakarla and his team for their help with the human
evaluation presented in this work.

\bibliography{anthology,custom}

\begin{thebibliography}{58}
\expandafter\ifx\csname natexlab\endcsname\relax\def\natexlab#1{#1}\fi

\bibitem[{Adlakha et~al.(2023)Adlakha, BehnamGhader, Lu, Meade, and Reddy}]{adlakhaetal}
Vaibhav Adlakha, Parishad BehnamGhader, Xing~Han Lu, Nicholas Meade, and Siva Reddy. 2023.
\newblock \href {https://doi.org/https://doi.org/10.48550/arXiv.2307.16877} {Evaluating correctness and faithfulness of instruction-following models for question answering}.
\newblock \emph{arXiv preprint arXiv:2307.16877}.

\bibitem[{Adolphs et~al.(2022)Adolphs, Shuster, Urbanek, Szlam, and Weston}]{adolphs2021reason}
Leonard Adolphs, Kurt Shuster, Jack Urbanek, Arthur Szlam, and Jason Weston. 2022.
\newblock \href {https://doi.org/10.18653/v1/2022.findings-emnlp.527} {Reason first, then respond: Modular generation for knowledge-infused dialogue}.
\newblock In \emph{Findings of the Association for Computational Linguistics: EMNLP 2022}, pages 7112--7132, Abu Dhabi, United Arab Emirates. Association for Computational Linguistics.

\bibitem[{Aksitov et~al.(2023)Aksitov, Chang, Reitter, Shakeri, and Sung}]{aksitov2023characterizing}
Renat Aksitov, Chung-Ching Chang, David Reitter, Siamak Shakeri, and Yunhsuan Sung. 2023.
\newblock \href {https://doi.org/https://doi.org/10.48550/arXiv.2302.05578} {Characterizing attribution and fluency tradeoffs for retrieval-augmented large language models}.
\newblock \emph{arXiv preprint arXiv:2302.05578}.

\bibitem[{Aralikatte et~al.(2021)Aralikatte, Narayan, Maynez, Rothe, and McDonald}]{aralikatte-etal-2021-focus}
Rahul Aralikatte, Shashi Narayan, Joshua Maynez, Sascha Rothe, and Ryan McDonald. 2021.
\newblock \href {https://doi.org/10.18653/v1/2021.acl-long.474} {Focus attention: Promoting faithfulness and diversity in summarization}.
\newblock In \emph{Proceedings of the 59th Annual Meeting of the Association for Computational Linguistics and the 11th International Joint Conference on Natural Language Processing (Volume 1: Long Papers)}, pages 6078--6095, Online. Association for Computational Linguistics.

\bibitem[{Baek et~al.(2023)Baek, Jeong, Kang, Park, and Hwang}]{baeketal}
Jinheon Baek, Soyeong Jeong, Minki Kang, Jong Park, and Sung Hwang. 2023.
\newblock \href {https://doi.org/10.18653/v1/2023.emnlp-main.107} {Knowledge-augmented language model verification}.
\newblock In \emph{Proceedings of the 2023 Conference on Empirical Methods in Natural Language Processing}, pages 1720--1736, Singapore. Association for Computational Linguistics.

\bibitem[{Chen et~al.(2023)Chen, Wu, Chen, and Chen}]{chen-etal-2023-fidelity}
Wei-Lin Chen, Cheng-Kuang Wu, Hsin-Hsi Chen, and Chung-Chi Chen. 2023.
\newblock \href {https://doi.org/10.18653/v1/2023.emnlp-main.54} {Fidelity-enriched contrastive search: Reconciling the faithfulness-diversity trade-off in text generation}.
\newblock In \emph{Proceedings of the 2023 Conference on Empirical Methods in Natural Language Processing}, pages 843--851, Singapore. Association for Computational Linguistics.

\bibitem[{Chiesurin et~al.(2023)Chiesurin, Dimakopoulos, Sobrevilla~Cabezudo, Eshghi, Papaioannou, Rieser, and Konstas}]{chiesurinetal}
Sabrina Chiesurin, Dimitris Dimakopoulos, Marco~Antonio Sobrevilla~Cabezudo, Arash Eshghi, Ioannis Papaioannou, Verena Rieser, and Ioannis Konstas. 2023.
\newblock \href {https://doi.org/10.18653/v1/2023.findings-acl.60} {The dangers of trusting stochastic parrots: Faithfulness and trust in open-domain conversational question answering}.
\newblock In \emph{Findings of the Association for Computational Linguistics: ACL 2023}, pages 947--959, Toronto, Canada. Association for Computational Linguistics.

\bibitem[{Cho and May(2020)}]{cho-may-2020-yesand}
Hyundong Cho and Jonathan May. 2020.
\newblock \href {https://doi.org/10.18653/v1/2020.acl-main.218} {Grounding conversations with improvised dialogues}.
\newblock In \emph{Proceedings of the 58th Annual Meeting of the Association for Computational Linguistics}, pages 2398--2413, Online. Association for Computational Linguistics.

\bibitem[{Daheim et~al.(2023)Daheim, Dziri, Sachan, Gurevych, and Ponti}]{daheim2023elastic}
Nico Daheim, Nouha Dziri, Mrinmaya Sachan, Iryna Gurevych, and Edoardo~M Ponti. 2023.
\newblock \href {https://doi.org/https://doi.org/10.48550/arXiv.2303.17574} {Elastic weight removal for faithful and abstractive dialogue generation}.
\newblock \emph{arXiv preprint arXiv:2303.17574}.

\bibitem[{Deng et~al.(2023)Deng, Zhang, Huang, and Hu}]{deng-etal-2023-towards}
Yifan Deng, Xingsheng Zhang, Heyan Huang, and Yue Hu. 2023.
\newblock \href {https://doi.org/10.18653/v1/2023.acl-long.250} {Towards faithful dialogues via focus learning}.
\newblock In \emph{Proceedings of the 61st Annual Meeting of the Association for Computational Linguistics (Volume 1: Long Papers)}, pages 4554--4566, Toronto, Canada. Association for Computational Linguistics.

\bibitem[{Dinan et~al.(2018)Dinan, Roller, Shuster, Fan, Auli, and Weston}]{dinan2018wizard}
Emily Dinan, Stephen Roller, Kurt Shuster, Angela Fan, Michael Auli, and Jason Weston. 2018.
\newblock \href {https://doi.org/https://doi.org/10.48550/arXiv.1811.01241} {Wizard of wikipedia: Knowledge-powered conversational agents}.
\newblock In \emph{Proceedings of International Conference on Learning Representations (ICLR)}.

\bibitem[{Dreyer et~al.(2023)Dreyer, Liu, Nan, Atluri, and Ravi}]{dreyer-etal-2023-evaluating}
Markus Dreyer, Mengwen Liu, Feng Nan, Sandeep Atluri, and Sujith Ravi. 2023.
\newblock \href {https://doi.org/10.18653/v1/2023.findings-eacl.156} {Evaluating the tradeoff between abstractiveness and factuality in abstractive summarization}.
\newblock In \emph{Findings of the Association for Computational Linguistics: EACL 2023}, pages 2089--2105, Dubrovnik, Croatia. Association for Computational Linguistics.

\bibitem[{Dziri et~al.(2022{\natexlab{a}})Dziri, Kamalloo, Milton, Zaiane, Yu, Ponti, and Reddy}]{dziri-etal-2022-faithdial}
Nouha Dziri, Ehsan Kamalloo, Sivan Milton, Osmar Zaiane, Mo~Yu, Edoardo~M. Ponti, and Siva Reddy. 2022{\natexlab{a}}.
\newblock \href {https://doi.org/10.1162/tacl_a_00529} {{F}aith{D}ial: A faithful benchmark for information-seeking dialogue}.
\newblock \emph{Transactions of the Association for Computational Linguistics}, 10:1473--1490.

\bibitem[{Dziri et~al.(2021)Dziri, Madotto, Za{\"\i}ane, and Bose}]{dziri-etal-2021-neural}
Nouha Dziri, Andrea Madotto, Osmar Za{\"\i}ane, and Avishek~Joey Bose. 2021.
\newblock \href {https://doi.org/10.18653/v1/2021.emnlp-main.168} {Neural path hunter: Reducing hallucination in dialogue systems via path grounding}.
\newblock In \emph{Proceedings of the 2021 Conference on Empirical Methods in Natural Language Processing}, pages 2197--2214, Online and Punta Cana, Dominican Republic. Association for Computational Linguistics.

\bibitem[{Dziri et~al.(2022{\natexlab{b}})Dziri, Milton, Yu, Zaiane, and Reddy}]{dziri2022origin}
Nouha Dziri, Sivan Milton, Mo~Yu, Osmar Zaiane, and Siva Reddy. 2022{\natexlab{b}}.
\newblock \href {https://doi.org/10.18653/v1/2022.naacl-main.387} {On the origin of hallucinations in conversational models: Is it the datasets or the models?}
\newblock pages 5271--5285.

\bibitem[{Ghazvininejad et~al.(2018)Ghazvininejad, Brockett, Chang, Dolan, Gao, Yih, and Galley}]{ghazvininejad2018knowledge}
Marjan Ghazvininejad, Chris Brockett, Ming-Wei Chang, Bill Dolan, Jianfeng Gao, Wen-tau Yih, and Michel Galley. 2018.
\newblock \href {https://doi.org/https://doi.org/10.1609/aaai.v32i1.11977} {A knowledge-grounded neural conversation model}.
\newblock In \emph{Proceedings of the Thirty-Second AAAI Conference on Artificial Intelligence and Thirtieth Innovative Applications of Artificial Intelligence Conference and Eighth AAAI Symposium on Educational Advances in Artificial Intelligence}, pages 5110--5117.

\bibitem[{Gopalakrishnan et~al.(2019)Gopalakrishnan, Hedayatnia, Chen, Gottardi, Kwatra, Venkatesh, Gabriel, and Hakkani-Tür}]{gopalakrishnan19_interspeech}
Karthik Gopalakrishnan, Behnam Hedayatnia, Qinlang Chen, Anna Gottardi, Sanjeev Kwatra, Anu Venkatesh, Raefer Gabriel, and Dilek Hakkani-Tür. 2019.
\newblock \href {https://doi.org/10.21437/Interspeech.2019-3079} {{Topical-Chat: Towards Knowledge-Grounded Open-Domain Conversations}}.
\newblock In \emph{Proc. Interspeech 2019}, pages 1891--1895.

\bibitem[{Gupta et~al.(2022)Gupta, Jhamtani, and Bigham}]{gupta-etal-2022-flow}
Prakhar Gupta, Harsh Jhamtani, and Jeffrey Bigham. 2022.
\newblock \href {https://doi.org/10.18653/v1/2022.findings-naacl.97} {Target-guided dialogue response generation using commonsense and data augmentation}.
\newblock In \emph{Findings of the Association for Computational Linguistics: NAACL 2022}, pages 1301--1317, Seattle, United States. Association for Computational Linguistics.

\bibitem[{Holtzman et~al.(2020)Holtzman, Buys, Forbes, and Choi}]{Holtzman2019TheCC}
Ari Holtzman, Jan Buys, Maxwell Forbes, and Yejin Choi. 2020.
\newblock \href {http://arxiv.org/abs/1904.09751} {The curious case of neural text degeneration}.
\newblock In \emph{Proceedings of International Conference on Learning Representations (ICLR)}.

\bibitem[{Honovich et~al.(2021)Honovich, Choshen, Aharoni, Neeman, Szpektor, and Abend}]{honovich-etal-2021-q2}
Or~Honovich, Leshem Choshen, Roee Aharoni, Ella Neeman, Idan Szpektor, and Omri Abend. 2021.
\newblock \href {https://doi.org/10.18653/v1/2021.emnlp-main.619} {$q^{2}$: {E}valuating factual consistency in knowledge-grounded dialogues via question generation and question answering}.
\newblock In \emph{Proceedings of the 2021 Conference on Empirical Methods in Natural Language Processing}, pages 7856--7870, Online and Punta Cana, Dominican Republic. Association for Computational Linguistics.

\bibitem[{Howcroft et~al.(2020)Howcroft, Belz, Clinciu, Gkatzia, Hasan, Mahamood, Mille, van Miltenburg, Santhanam, and Rieser}]{howcroft-etal-2020-twenty}
David~M. Howcroft, Anya Belz, Miruna-Adriana Clinciu, Dimitra Gkatzia, Sadid~A. Hasan, Saad Mahamood, Simon Mille, Emiel van Miltenburg, Sashank Santhanam, and Verena Rieser. 2020.
\newblock \href {https://aclanthology.org/2020.inlg-1.23} {Twenty years of confusion in human evaluation: {NLG} needs evaluation sheets and standardised definitions}.
\newblock In \emph{Proceedings of the 13th International Conference on Natural Language Generation}, pages 169--182, Dublin, Ireland. Association for Computational Linguistics.

\bibitem[{Hu et~al.(2022)Hu, Chan, Liu, Xiao, Wu, and Huang}]{hu-etal-2022-planet}
Zhe Hu, Hou~Pong Chan, Jiachen Liu, Xinyan Xiao, Hua Wu, and Lifu Huang. 2022.
\newblock \href {https://doi.org/10.18653/v1/2022.acl-long.163} {{PLANET}: Dynamic content planning in autoregressive transformers for long-form text generation}.
\newblock In \emph{Proceedings of the 60th Annual Meeting of the Association for Computational Linguistics (Volume 1: Long Papers)}, pages 2288--2305, Dublin, Ireland. Association for Computational Linguistics.

\bibitem[{Hua and Wang(2019)}]{hua-wang-2019-sentence}
Xinyu Hua and Lu~Wang. 2019.
\newblock \href {https://doi.org/10.18653/v1/D19-1055} {Sentence-level content planning and style specification for neural text generation}.
\newblock In \emph{Proceedings of the 2019 Conference on Empirical Methods in Natural Language Processing and the 9th International Joint Conference on Natural Language Processing (EMNLP-IJCNLP)}, pages 591--602, Hong Kong, China. Association for Computational Linguistics.

\bibitem[{Jang et~al.(2023)Jang, Son, Lee, Son, Hur, Lim, Moon, Yang, and Lim}]{jang-etal-2023-post}
Yoonna Jang, Suhyune Son, Jeongwoo Lee, Junyoung Son, Yuna Hur, Jungwoo Lim, Hyeonseok Moon, Kisu Yang, and Heuiseok Lim. 2023.
\newblock \href {https://doi.org/10.18653/v1/2023.emnlp-main.295} {Post-hoc utterance refining method by entity mining for faithful knowledge grounded conversations}.
\newblock In \emph{Proceedings of the 2023 Conference on Empirical Methods in Natural Language Processing}, pages 4844--4861, Singapore. Association for Computational Linguistics.

\bibitem[{Kim et~al.(2022)Kim, Ahn, Kim, and Lee}]{kim-etal-2022-flow}
Wongyu Kim, Youbin Ahn, Donghyun Kim, and Kyong-Ho Lee. 2022.
\newblock \href {https://doi.org/10.18653/v1/2022.naacl-main.303} {Emp-{RFT}: Empathetic response generation via recognizing feature transitions between utterances}.
\newblock In \emph{Proceedings of the 2022 Conference of the North American Chapter of the Association for Computational Linguistics: Human Language Technologies}, pages 4118--4128, Seattle, United States. Association for Computational Linguistics.

\bibitem[{Kodama et~al.(2023)Kodama, Kiyomaru, Huang, Okahisa, and Kurohashi}]{kodama-etal-2023-knowledge}
Takashi Kodama, Hirokazu Kiyomaru, Yin~Jou Huang, Taro Okahisa, and Sadao Kurohashi. 2023.
\newblock \href {https://doi.org/10.18653/v1/2023.acl-srw.34} {Is a knowledge-based response engaging?: An analysis on knowledge-grounded dialogue with information source annotation}.
\newblock In \emph{Proceedings of the 61st Annual Meeting of the Association for Computational Linguistics (Volume 4: Student Research Workshop)}, pages 237--243, Toronto, Canada. Association for Computational Linguistics.

\bibitem[{Ladhak et~al.(2022)Ladhak, Durmus, He, Cardie, and McKeown}]{ladhak-etal-2022-faithful}
Faisal Ladhak, Esin Durmus, He~He, Claire Cardie, and Kathleen McKeown. 2022.
\newblock \href {https://doi.org/10.18653/v1/2022.acl-long.100} {Faithful or extractive? on mitigating the faithfulness-abstractiveness trade-off in abstractive summarization}.
\newblock In \emph{Proceedings of the 60th Annual Meeting of the Association for Computational Linguistics (Volume 1: Long Papers)}, pages 1410--1421, Dublin, Ireland. Association for Computational Linguistics.

\bibitem[{Li et~al.(2017)Li, Su, Shen, Li, Cao, and Niu}]{li-etal-2017-dailydialog}
Yanran Li, Hui Su, Xiaoyu Shen, Wenjie Li, Ziqiang Cao, and Shuzi Niu. 2017.
\newblock \href {https://aclanthology.org/I17-1099} {{D}aily{D}ialog: A manually labelled multi-turn dialogue dataset}.
\newblock In \emph{Proceedings of the Eighth International Joint Conference on Natural Language Processing (Volume 1: Long Papers)}, pages 986--995, Taipei, Taiwan. Asian Federation of Natural Language Processing.

\bibitem[{Liu et~al.(2019)Liu, Ott, Goyal, Du, Joshi, Chen, Levy, Lewis, Zettlemoyer, and Stoyanov}]{roberta}
Yinhan Liu, Myle Ott, Naman Goyal, Jingfei Du, Mandar Joshi, Danqi Chen, Omer Levy, M.~Lewis, Luke Zettlemoyer, and Veselin Stoyanov. 2019.
\newblock \href {https://doi.org/https://doi.org/10.48550/arXiv.1907.11692} {Roberta: A robustly optimized bert pretraining approach}.
\newblock \emph{ArXiv}, abs/1907.11692.

\bibitem[{Mallinson et~al.(2020)Mallinson, Severyn, Malmi, and Garrido}]{mallinson2020felix}
Jonathan Mallinson, Aliaksei Severyn, Eric Malmi, and Guillermo Garrido. 2020.
\newblock \href {https://doi.org/10.18653/v1/2020.findings-emnlp.111} {{FELIX}: Flexible text editing through tagging and insertion}.
\newblock In \emph{Findings of the Association for Computational Linguistics: EMNLP 2020}, pages 1244--1255, Online. Association for Computational Linguistics.

\bibitem[{Malmi et~al.(2020)Malmi, Severyn, and Rothe}]{malmi2020unsupervised}
Eric Malmi, Aliaksei Severyn, and Sascha Rothe. 2020.
\newblock \href {https://doi.org/10.18653/v1/2020.emnlp-main.699} {Unsupervised text style transfer with padded masked language models}.
\newblock In \emph{Proceedings of the 2020 Conference on Empirical Methods in Natural Language Processing (EMNLP)}, pages 8671--8680.

\bibitem[{Nandwani et~al.(2023)Nandwani, Kumar, Raghu, Joshi, and Lastras}]{nandwani-etal-2023-pointwise}
Yatin Nandwani, Vineet Kumar, Dinesh Raghu, Sachindra Joshi, and Luis Lastras. 2023.
\newblock \href {https://doi.org/10.18653/v1/2023.emnlp-main.639} {Pointwise mutual information based metric and decoding strategy for faithful generation in document grounded dialogs}.
\newblock In \emph{Proceedings of the 2023 Conference on Empirical Methods in Natural Language Processing}, pages 10335--10347, Singapore. Association for Computational Linguistics.

\bibitem[{Narayan et~al.(2023)Narayan, Maynez, Amplayo, Ganchev, Louis, Huot, Sandholm, Das, and Lapata}]{narayan2022conditional}
Shashi Narayan, Joshua Maynez, Reinald~Kim Amplayo, Kuzman Ganchev, Annie Louis, Fantine Huot, Anders Sandholm, Dipanjan Das, and Mirella Lapata. 2023.
\newblock \href {https://doi.org/10.1162/tacl_a_00583} {Conditional generation with a question-answering blueprint}.
\newblock \emph{Transactions of the Association for Computational Linguistics}, 11:974--996.

\bibitem[{Narayan et~al.(2021)Narayan, Zhao, Maynez, Sim{\~o}es, Nikolaev, and McDonald}]{narayan2021planning}
Shashi Narayan, Yao Zhao, Joshua Maynez, Gon{\c{c}}alo Sim{\~o}es, Vitaly Nikolaev, and Ryan McDonald. 2021.
\newblock \href {https://doi.org/10.1162/tacl_a_00438} {Planning with learned entity prompts for abstractive summarization}.
\newblock \emph{Transactions of the Association for Computational Linguistics}, 9:1475--1492.

\bibitem[{Peng et~al.(2023)Peng, Galley, He, Cheng, Xie, Hu, Huang, Liden, Yu, Chen et~al.}]{pengetalcritique}
Baolin Peng, Michel Galley, Pengcheng He, Hao Cheng, Yujia Xie, Yu~Hu, Qiuyuan Huang, Lars Liden, Zhou Yu, Weizhu Chen, et~al. 2023.
\newblock \href {https://doi.org/https://doi.org/10.48550/arXiv.2302.12813} {Check your facts and try again: Improving large language models with external knowledge and automated feedback}.
\newblock \emph{arXiv preprint arXiv:2302.12813}.

\bibitem[{Raffel et~al.(2020)Raffel, Shazeer, Roberts, Lee, Narang, Matena, Zhou, Li, and Liu}]{raffel2020exploring}
Colin Raffel, Noam Shazeer, Adam Roberts, Katherine Lee, Sharan Narang, Michael Matena, Yanqi Zhou, Wei Li, and Peter~J. Liu. 2020.
\newblock \href {http://jmlr.org/papers/v21/20-074.html} {Exploring the limits of transfer learning with a unified text-to-text transformer}.
\newblock \emph{Journal of Machine Learning Research}, 21(140):1--67.

\bibitem[{Rashkin et~al.(2020)Rashkin, Celikyilmaz, Choi, and Gao}]{rashkin2020plotmachines}
Hannah Rashkin, Asli Celikyilmaz, Yejin Choi, and Jianfeng Gao. 2020.
\newblock \href {https://doi.org/10.18653/v1/2020.emnlp-main.349} {Plotmachines: Outline-conditioned generation with dynamic plot state tracking}.
\newblock In \emph{Proceedings of the 2020 Conference on Empirical Methods in Natural Language Processing (EMNLP)}, pages 4274--4295.

\bibitem[{Rashkin et~al.(2023)Rashkin, Nikolaev, Lamm, Aroyo, Collins, Das, Petrov, Tomar, Turc, and Reitter}]{rashkin-etal-2023-ais}
Hannah Rashkin, Vitaly Nikolaev, Matthew Lamm, Lora Aroyo, Michael Collins, Dipanjan Das, Slav Petrov, Gaurav~Singh Tomar, Iulia Turc, and David Reitter. 2023.
\newblock \href {https://doi.org/10.1162/coli_a_00486} {{Measuring Attribution in Natural Language Generation Models}}.
\newblock \emph{Computational Linguistics}, pages 1--64.

\bibitem[{Rashkin et~al.(2021)Rashkin, Reitter, Tomar, and Das}]{rashkin-etal-2021-increasing}
Hannah Rashkin, David Reitter, Gaurav~Singh Tomar, and Dipanjan Das. 2021.
\newblock \href {https://doi.org/10.18653/v1/2021.acl-long.58} {Increasing faithfulness in knowledge-grounded dialogue with controllable features}.
\newblock In \emph{Proceedings of the 59th Annual Meeting of the Association for Computational Linguistics and the 11th International Joint Conference on Natural Language Processing (Volume 1: Long Papers)}, pages 704--718, Online. Association for Computational Linguistics.

\bibitem[{Sevegnani et~al.(2021)Sevegnani, Howcroft, Konstas, and Rieser}]{sevegnani-etal-2021-flow}
Karin Sevegnani, David~M. Howcroft, Ioannis Konstas, and Verena Rieser. 2021.
\newblock \href {https://doi.org/10.18653/v1/2021.acl-long.194} {{OTT}ers: One-turn topic transitions for open-domain dialogue}.
\newblock In \emph{Proceedings of the 59th Annual Meeting of the Association for Computational Linguistics and the 11th International Joint Conference on Natural Language Processing (Volume 1: Long Papers)}, pages 2492--2504, Online. Association for Computational Linguistics.

\bibitem[{Shuster et~al.(2020)Shuster, Ju, Roller, Dinan, Boureau, and Weston}]{shuster2020dialogue}
Kurt Shuster, Da~Ju, Stephen Roller, Emily Dinan, Y-Lan Boureau, and Jason Weston. 2020.
\newblock \href {https://doi.org/10.18653/v1/2020.acl-main.222} {The dialogue dodecathlon: Open-domain knowledge and image grounded conversational agents}.
\newblock In \emph{Proceedings of the 58th Annual Meeting of the Association for Computational Linguistics}, pages 2453--2470.

\bibitem[{Sun et~al.(2023)Sun, Li, Mi, Bie, Li, and Li}]{sun-etal-2023-towards}
Bin Sun, Yitong Li, Fei Mi, Fanhu Bie, Yiwei Li, and Kan Li. 2023.
\newblock \href {https://doi.org/10.18653/v1/2023.acl-short.148} {Towards fewer hallucinations in knowledge-grounded dialogue generation via augmentative and contrastive knowledge-dialogue}.
\newblock In \emph{Proceedings of the 61st Annual Meeting of the Association for Computational Linguistics (Volume 2: Short Papers)}, pages 1741--1750, Toronto, Canada. Association for Computational Linguistics.

\bibitem[{Tan et~al.(2021)Tan, Yang, Al-Shedivat, Xing, and Hu}]{tan-etal-2021}
Bowen Tan, Zichao Yang, Maruan Al-Shedivat, Eric Xing, and Zhiting Hu. 2021.
\newblock \href {https://doi.org/10.18653/v1/2021.naacl-main.341} {Progressive generation of long text with pretrained language models}.
\newblock In \emph{Proceedings of the 2021 Conference of the North American Chapter of the Association for Computational Linguistics: Human Language Technologies}, pages 4313--4324, Online. Association for Computational Linguistics.

\bibitem[{Thoppilan et~al.(2022)Thoppilan, Freitas, Hall, Shazeer, Kulshreshtha, Cheng, Jin, Bos, Baker, Du, Li, Lee, Zheng, Ghafouri, Menegali, Huang, Krikun, Lepikhin, Qin, Chen, Xu, Chen, Roberts, Bosma, Zhao, Zhou, Chang, Krivokon, Rusch, Pickett, Srinivasan, Man, Meier-Hellstern, Morris, Doshi, Santos, Duke, Soraker, Zevenbergen, Prabhakaran, Diaz, Hutchinson, Olson, Molina, Hoffman-John, Lee, Aroyo, Rajakumar, Butryna, Lamm, Kuzmina, Fenton, Cohen, Bernstein, Kurzweil, Aguera-Arcas, Cui, Croak, Chi, and Le}]{lamda}
Romal Thoppilan, Daniel~De Freitas, Jamie Hall, Noam Shazeer, Apoorv Kulshreshtha, Heng-Tze Cheng, Alicia Jin, Taylor Bos, Leslie Baker, Yu~Du, YaGuang Li, Hongrae Lee, Huaixiu~Steven Zheng, Amin Ghafouri, Marcelo Menegali, Yanping Huang, Maxim Krikun, Dmitry Lepikhin, James Qin, Dehao Chen, Yuanzhong Xu, Zhifeng Chen, Adam Roberts, Maarten Bosma, Vincent Zhao, Yanqi Zhou, Chung-Ching Chang, Igor Krivokon, Will Rusch, Marc Pickett, Pranesh Srinivasan, Laichee Man, Kathleen Meier-Hellstern, Meredith~Ringel Morris, Tulsee Doshi, Renelito~Delos Santos, Toju Duke, Johnny Soraker, Ben Zevenbergen, Vinodkumar Prabhakaran, Mark Diaz, Ben Hutchinson, Kristen Olson, Alejandra Molina, Erin Hoffman-John, Josh Lee, Lora Aroyo, Ravi Rajakumar, Alena Butryna, Matthew Lamm, Viktoriya Kuzmina, Joe Fenton, Aaron Cohen, Rachel Bernstein, Ray Kurzweil, Blaise Aguera-Arcas, Claire Cui, Marian Croak, Ed~Chi, and Quoc Le. 2022.
\newblock \href {https://doi.org/https://doi.org/10.48550/arXiv.2201.08239} {Lamda: Language models for dialog applications}.
\newblock \emph{arXiv preprint arXiv:2201.08239}.

\bibitem[{Tian and Peng(2022)}]{tian2022sonnet}
Yufei Tian and Nanyun Peng. 2022.
\newblock \href {https://doi.org/10.18653/v1/2022.naacl-main.262} {Zero-shot sonnet generation with discourse-level planning and aesthetics features}.
\newblock In \emph{2022 Annual Conference of the North American Chapter of the Association for Computational Linguistics (NAACL)}, pages 3587--3597.

\bibitem[{Tian et~al.(2020)Tian, Bi, Lee, Xue, Song, Liu, and Zhang}]{tian2020response}
Zhiliang Tian, Wei Bi, Dongkyu Lee, Lanqing Xue, Yiping Song, Xiaojiang Liu, and Nevin~Lianwen Zhang. 2020.
\newblock \href {https://doi.org/10.18653/v1/2020.acl-main.61} {Response-anticipated memory for on-demand knowledge integration in response generation}.
\newblock In \emph{Proceedings of the 58th Annual Meeting of the Association for Computational Linguistics}, pages 650--659.

\bibitem[{Vaswani et~al.(2017)Vaswani, Shazeer, Parmar, Uszkoreit, Jones, Gomez, Kaiser, and Polosukhin}]{vaswani2017attention}
Ashish Vaswani, Noam Shazeer, Niki Parmar, Jakob Uszkoreit, Llion Jones, Aidan~N Gomez, {\L}ukasz Kaiser, and Illia Polosukhin. 2017.
\newblock \href {https://proceedings.neurips.cc/paper_files/paper/2017/file/3f5ee243547dee91fbd053c1c4a845aa-Paper.pdf} {Attention is all you need}.
\newblock \emph{Advances in neural information processing systems}, 30.

\bibitem[{Wang et~al.(2023{\natexlab{a}})Wang, Wang, Mi, Deng, Wang, Liang, Xu, and Wong}]{wangetalcot}
Hongru Wang, Rui Wang, Fei Mi, Yang Deng, Zezhong Wang, Bin Liang, Ruifeng Xu, and Kam-Fai Wong. 2023{\natexlab{a}}.
\newblock \href {https://doi.org/10.18653/v1/2023.findings-emnlp.806} {Cue-{C}o{T}: Chain-of-thought prompting for responding to in-depth dialogue questions with {LLM}s}.
\newblock In \emph{Findings of the Association for Computational Linguistics: EMNLP 2023}, pages 12047--12064, Singapore. Association for Computational Linguistics.

\bibitem[{Wang et~al.(2023{\natexlab{b}})Wang, Ding, Cao, Tian, Wang, Tao, and Guo}]{wangetalforgetting}
Qingyue Wang, Liang Ding, Yanan Cao, Zhiliang Tian, Shi Wang, Dacheng Tao, and Li~Guo. 2023{\natexlab{b}}.
\newblock \href {https://doi.org/https://doi.org/10.48550/arXiv.2308.15022} {Recursively summarizing enables long-term dialogue memory in large language models}.
\newblock \emph{arXiv preprint arXiv:2308.15022}.

\bibitem[{Wilie et~al.(2023)Wilie, Xu, Chung, Cahyawijaya, Lovenia, and Fung}]{wilie2023pick}
Bryan Wilie, Yan Xu, Willy Chung, Samuel Cahyawijaya, Holy Lovenia, and Pascale Fung. 2023.
\newblock \href {https://doi.org/https://doi.org/10.48550/arXiv.2309.10413} {Pick: Polished \& informed candidate scoring for knowledge-grounded dialogue systems}.
\newblock \emph{arXiv preprint arXiv:2309.10413}.

\bibitem[{Williams et~al.(2018)Williams, Nangia, and Bowman}]{mnli}
Adina Williams, Nikita Nangia, and Samuel Bowman. 2018.
\newblock \href {http://aclweb.org/anthology/N18-1101} {A broad-coverage challenge corpus for sentence understanding through inference}.
\newblock In \emph{Proceedings of the 2018 Conference of the North American Chapter of the Association for Computational Linguistics: Human Language Technologies, Volume 1 (Long Papers)}, pages 1112--1122. Association for Computational Linguistics.

\bibitem[{Yao et~al.(2019)Yao, Peng, Weischedel, Knight, Zhao, and Yan}]{yao-etal-2019-pnw}
Lili Yao, Nanyun Peng, Ralph Weischedel, Kevin Knight, Dongyan Zhao, and Rui Yan. 2019.
\newblock \href {https://doi.org/10.1609/aaai.v33i01.33017378} {Plan-and-write: Towards better automatic storytelling}.
\newblock In \emph{Proceedings of the Thirty-Third AAAI Conference on Artificial Intelligence and Thirty-First Innovative Applications of Artificial Intelligence Conference and Ninth AAAI Symposium on Educational Advances in Artificial Intelligence}, AAAI'19/IAAI'19/EAAI'19. AAAI Press.

\bibitem[{Zhang et~al.(2020)Zhang, Sun, Galley, Chen, Brockett, Gao, Gao, Liu, and Dolan}]{zhang2019dialogpt}
Yizhe Zhang, Siqi Sun, Michel Galley, Yen-Chun Chen, Chris Brockett, Xiang Gao, Jianfeng Gao, Jingjing Liu, and Bill Dolan. 2020.
\newblock \href {https://doi.org/10.18653/v1/2020.acl-demos.30} {{DIALOGPT} : Large-scale generative pre-training for conversational response generation}.
\newblock In \emph{Proceedings of the 58th Annual Meeting of the Association for Computational Linguistics: System Demonstrations}, pages 270--278, Online. Association for Computational Linguistics.

\bibitem[{Zheng et~al.(2022)Zheng, Wang, Ke, Yang, and Huang}]{wu2021semantic}
Yinhe Zheng, Yida Wang, Pei Ke, Zhenyu Yang, and Minlie Huang. 2022.
\newblock \href {http://arxiv.org/abs/2106.03065} {Semantic-enhanced explainable finetuning for open-domain dialogues}.

\bibitem[{Zhou et~al.(2018)Zhou, Prabhumoye, and Black}]{zhou-etal-2018-dataset}
Kangyan Zhou, Shrimai Prabhumoye, and Alan~W Black. 2018.
\newblock \href {https://doi.org/10.18653/v1/D18-1076} {A dataset for document grounded conversations}.
\newblock In \emph{Proceedings of the 2018 Conference on Empirical Methods in Natural Language Processing}, pages 708--713, Brussels, Belgium. Association for Computational Linguistics.

\bibitem[{Zhou et~al.(2022{\natexlab{a}})Zhou, Cho, Jandaghi, Lee, Lin, Pujara, and Ren}]{zhou-etal-2022-reflect}
Pei Zhou, Hyundong Cho, Pegah Jandaghi, Dong-Ho Lee, Bill~Yuchen Lin, Jay Pujara, and Xiang Ren. 2022{\natexlab{a}}.
\newblock \href {https://doi.org/10.18653/v1/2022.emnlp-main.714} {Reflect, not reflex: Inference-based common ground improves dialogue response quality}.
\newblock In \emph{Proceedings of the 2022 Conference on Empirical Methods in Natural Language Processing}, pages 10450--10468, Abu Dhabi, United Arab Emirates. Association for Computational Linguistics.

\bibitem[{Zhou et~al.(2022{\natexlab{b}})Zhou, Gopalakrishnan, Hedayatnia, Kim, Pujara, Ren, Liu, and Hakkani-Tur}]{zhou-etal-2022-think}
Pei Zhou, Karthik Gopalakrishnan, Behnam Hedayatnia, Seokhwan Kim, Jay Pujara, Xiang Ren, Yang Liu, and Dilek Hakkani-Tur. 2022{\natexlab{b}}.
\newblock \href {https://doi.org/10.18653/v1/2022.acl-long.88} {Think before you speak: Explicitly generating implicit commonsense knowledge for response generation}.
\newblock In \emph{Proceedings of the 60th Annual Meeting of the Association for Computational Linguistics (Volume 1: Long Papers)}, pages 1237--1252, Dublin, Ireland. Association for Computational Linguistics.

\bibitem[{Zou et~al.(2021)Zou, Liu, Hu, and Zhang}]{zou2021thinking}
Yicheng Zou, Zhihua Liu, Xingwu Hu, and Qi~Zhang. 2021.
\newblock \href {https://doi.org/10.18653/v1/2021.emnlp-main.169} {Thinking clearly, talking fast: Concept-guided non-autoregressive generation for open-domain dialogue systems}.
\newblock In \emph{Proceedings of the 2021 Conference on Empirical Methods in Natural Language Processing}, pages 2215--2226.

\end{thebibliography}
\bibstyle{acl_natbib}

\appendix

\section{Structural Variables}
\label{sec:appendix:struct}
Below, we describe each of the structural variables used in the \texttt{struct} and \texttt{full} plans:
\begin{itemize}
    \item dialogue acts -- labelled using a T5 classifier that was finetuned on DailyDialog chit-chat dataset \cite{li-etal-2017-dailydialog}
    \item emotion -- labelled using a T5 classifier that was trained on DailyDialog chit-chat dataset \cite{li-etal-2017-dailydialog}
    \item objective/personal voice -- using lexical matching to find instances of first person (see \citep{rashkin-etal-2021-increasing})
    \item linguistic specificity -- using idf scores of the output relative to the entire training set, split into high/med/low terciles
    \item nli score with evidence -- using nli classifier to find similarity to the evidence, split into entail/not-entail scores (see \citep{rashkin-etal-2021-increasing})
    \item lexical precision similarity with evidence -- precision score using lexical matching to find similarity to the evidence, split into high/med/low terciles (see \citep{rashkin-etal-2021-increasing})
    \item similarity (lexical precision) with previous turn by the apprentice --  precision score using lexical matching to find similarity of response to the previous apprentice turn (turn $i-1$), split into high/med/low terciles
    \item similarity (lexical precision) with previous turn by the wizard --  precision score using lexical matching to find similarity of response to the previous wizard turn (turn $i-2$), split into high/med/low terciles
\end{itemize}

\section{Data Examples}
\label{sec:data-examples-appendix}

In Table~\ref{tab:dataexamples}, we include gold examples from WoW training set with the constructed keyword and structural plans.

\subsection{Model Input and Output formatting}
\label{sec:io-format-appendix}

For the generation {\bf model G  input}, we use the format of:
"{\it the previous apprentice turn} [special-delimiter-1] {\it evidence} {\it and remaining conversation history in reverse order with delimiters separating conversation turns} [special-delimiter-2]

\noindent For the generation {\bf model G  output}, we use the format of:"{\it structural plan token sequence} [special-delimiter-3] {\it keyword plan token sequence} [special-delimiter-4]{\it generated response}."

So, for instance, in the second example from Table~\ref{tab:dataexamples}, this gets encoded as:

Input string: \texttt{ all of the nordic places in the netherlands seem really awesome and beautiful [special-delimiter-1] the southernmost of the scandinavian nations, it is south-west of sweden and south of norway, and bordered to the south by germany. [delimiter-wizard-turn] it  probably is! it’s   actually a kingdom, and is  nordic. it  is  a sovereign nation. [delimiter-apprentice-turn] denmark seems like a really cool place to visit
[special-delimiter-2]}
\\
Output string: \texttt{[dact:inform] [emo:neutral] [objective] [spec:med] [entail] [evidsim:high] [prevappsim:high] [prevwizsim:high]   [special-delimiter-3] denmark edge sweden norway germany [special-delimiter-4]  denmark is on the edge of sweden and norway and germany.}

\begin{table*}
    \centering
    \footnotesize
    \begin{tabular}{p{3cm}p{2.5cm}p{2.75cm}|p{3.65cm}|p{2.cm}}
    \toprule
    Conv. History& Evidence&Gold Response&Structural Plan & Keyword Plan\\
    \midrule
{\bf Wiz:}"i think science fiction is an amazing genre for anything. future science, technology, time travel, ftl travel, they're all such interesting concepts."\newline
{\bf App:} "i'm a huge fan of science fiction myself! "& science fiction films have often been used to focus on political or social issues, and to explore philosophical issues like the human condition. & awesome! i really love how sci-fi storytellers focus on political / social / philosophical issues that would still be around even in the future. makes them relatable. & \texttt{[dact:inform] [emo:neutral] [objective] [spec:high] [noentail] [evidsim:low] [prevappsim:med] [prevwizsim:med] }& storytellers issues future \\ \midrule
{\bf App:} "denmark seems like a really cool place to visit"\newline
{\bf Wiz:} "it probably is! it's actually a kingdom, and is nordic. it is a sovereign nation."\newline
{\bf App:} "all of the nordic places in the netherlands seem really awesome and beautiful." &
the southernmost of the scandinavian nations, it is south-west of sweden and south of norway, and bordered to the south by germany. & denmark is on the edge of sweden and norway and germany.
&  \texttt{[dact:inform] [emo:neutral] [objective] [spec:med] [entail] [evidsim:high] [prevappsim:high] [prevwizsim:high] }
& denmark edge sweden norway germany \\
\midrule
{\bf App:} "do you like cheeseburgers? they seem to be as popular now than they ever were." &traditionally, the slice of cheese is placed on top of the meat patty, but the burger can include many variations in structure, ingredients, and composition. & i love a huge dressed cheeseburger. traditionally the cheese is put on top of the patty but there are many variations.
&  \texttt{[dact:inform] [emo:happy] [personal] [spec:med] [noentail] [evidsim:high]  [prevappsim:low] [prevwizsim:low] }
& dressed cheeseburger cheese top patty variations \\
\midrule
{\bf Wiz:} "i've lined in new york city all my life. it's the best city on earth."\newline
{\bf App:} "how many people live in new york? "&with an estimated 2016 population of 8,537,673 distributed over a land area of about , new york city is also the most densely populated major city in the united states.& a few... 8,537,673 to be exact but some day's it feels like more. have you ever came to the city? 
&\texttt{ [dact:question] [emo:neutral] [objective] [spec:low] [noentail] [evidsim:low] [prevappsim:low] [prevwizsim:med]}
& day city \\

\bottomrule
    \end{tabular}
    \caption{Training Data Examples: examples from WoW training set with the heuristically constructed structural and keyword plan }
    \label{tab:dataexamples}
\end{table*}

\section{Plan Editor Model}
\label{sec:plan-editor-appendix}
We provide more details about the training and inference for the plan editor model below. These are based on the MASKER approach described in ~\citet{malmi2020unsupervised}.

\noindent\textbf{Training}: MASKER \cite{malmi2020unsupervised} is a non-autoregressive Roberta-style language model~\cite{roberta} using the Padded Masked Language Modeling (MLM) strategy~\cite{mallinson2020felix}. Padded MLM modifies the original MLM objective to also take into account the length of infilled tokens. Instead of masking a single token, this approach masks out a sequence of whole words up to $n_p$ tokens, filling the remaining tokens with [PAD] to ensure that the input always consists of $n_p$ [MASK] tokens. Then, the model is trained on the pseudo-likelihood of the original tokens $C_{i:j}$:
\begin{multline}
    L(C_{i:j} | C_{\textbackslash i:j};\Theta) = \prod_{t=i}^{j}P_{MLM}(c_t | C_{\textrm{\textbackslash} i:j};\Theta)\\ \times \prod_{t=j+1}^{i+n_p-1}P_{MLM}([PAD]_t | C_{\textrm{\textbackslash} i:j};\Theta) 
\end{multline}
$C_{i:j}$ denotes the full content plan without padding and where $C_{\textrm{\textbackslash} i:j}$ denotes the content plan with tokens $c_i...c_j$ masked out. $P_{MLM}(c_t | C_{\textrm{\textbackslash} i:j};\Theta)$ is the probability of the random variable corresponding to the t-th token in $C_{\textrm{\textbackslash} i:j}$ taking the value $c_t$ or [PAD]. Finally, $\Theta$ corresponds to either $\Theta_{\textrm{source}}$ or $\Theta_{\textrm{target}}$, depending on the data the model is trained on. In practice, a single unified model is trained by using a special indicator token [SOURCE] or [TARGET] in the input.  

\noindent\textbf{Inference}: For inference, the editor model needs to find a text span where the source and the target models disagree the most and then replace this with the maximum likelihood replacement suggested by the target model $\hat{C_{i:j}}^{\textrm{target}}$. Since the content plans are relatively shorter than entire utterances and bounded, we simply try out all the possible masking positions $i:j$ in order to maximize the score $S(i,j)$:

%, while still keeping the total masked length to $n_p$ using [PAD] tokens. The final position is decided by the score $S(i,j)$: 

\begin{equation}
    S(i,j) = TS(i,j) + SS(i,j),
\end{equation}
\begin{multline}
TS(i,j) = L(\hat{C_{i:j}}^{\textrm{target}} | C_{\textrm{\textbackslash} i:j};\Theta_{\textrm{target}}) \\ - L(C_{i:j} | C_{\textrm{\textbackslash} i:j};\Theta_{\textrm{target}})
\end{multline}
\begin{multline}
SS(i,j) = -\textrm{max}[0, L(\hat{C_{i:j}}^{\textrm{target}} | C_{\textrm{\textbackslash} i:j};\Theta_{\textrm{source}}) \\ - L(C_{i:j} | C_{\textrm{\textbackslash} i:j};\Theta_{\textrm{source}})]
\end{multline}

$TS(i,j)$ is the score computed with respect to the target model. Intuitively, a position is preferable if a) a good replacement is available and b) the existing tokens in this position are less likely under the target model.

The term $SS(i,j)$ evaluates $\hat{C_{i:j}}^{\textrm{target}}$ and $C_{i:j}$ under the source model to ensure that the edit is improving only in a way that improves in a way that affects the differences between target and source domain. Without this term, it is possible that the target model would want to make other changes to the content plan, such as replacing rare tokens with more common ones, which may not necessarily be related to the differences between the source and target domains.

\section{Plan Editing Examples}
\label{sec:plan-editing-examples-appendix}

In Table~\ref{tab:planeditexamples}, we show the inputs and outputs of the plan editing module for one example over multiple metric-aware editing steps.  Many of the updates to the structural attributes reflect that the model learns to increase attribution scores by gradually shifting the plan towards the third person, setting the entail variable to true, and increasing the lexical precision with the evidence.  

The output of the generation model using the original plan was ``i'm not sure, but i do know that iguanas can range in length, including their tail.''  After using metric-aware editing, the output of the generation model is ``yes, they can range in length, including their tail.''  We note that the output of the model using metric-aware editing is shorter and sticks more closely to words from the evidence, which likely means that it scores higher on our automatic metrics.  However, qualitatively, the output from using the metric-agnostic plan is a more apt response.

\begin{table*}[]
\centering
\begin{tabular}{p{2cm}p{3cm}|c|p{5cm}}
  evidence & conv history & edit timestep  & plan  \\
  \midrule
 \multirow{5}{2cm}{ iguanas can range from in length, including their tail.} &
 \multirow{10}{3cm}{
  i love iguanas, i have a few as pets. do you like lizards at all? \newline yes, i like them. they are interesting.and prehistoric looking. i like turtles too. \newline i agree, they definitely have a prehistoric look to them. there are also over 6000 species spread across the world. \newline do they have teeth and does their bite hurt if they bite you? }
 &0& [dact:inform] [emo:neutral] [personal] [spec:low] [nonentail]  [evidsim:low] [prevappsim:med] [prevwiz:high]  [special-delimiter-3] tail iguanas \\\cmidrule{3-4}
 &&1&  [dact:inform] [emo:neutral] [personal] [spec:low] [nonentail]  {\color{blue}[evidsim:med]} [prevappsim:med]  [prevwiz:high]  [special-delimiter-3] tail iguanas \\\cmidrule{3-4}
 &&2&[dact:inform] [emo:neutral] [personal] [spec:low] [nonentail] [evidsim:med] [prevappsim:med]  [prevwiz:high]  [special-delimiter-3] {\color{cyan}length} tail iguanas \\\cmidrule{3-4}
 &&3&[dact:inform] [emo:neutral]  {\color{blue}[objective]}  [spec:low] {\color{blue}[entail]} {\color{blue}[evidsim:high]} [prevappsim:med]  [prevwiz:high]  [special-delimiter-3] length tail {\color{red}iguanas}\\\cmidrule{3-4}
 &&...\\\cmidrule{3-4}
 &&9&  [dact:inform] [emo:neutral] [objective]   [spec:low]  [entail] [evidsim:high]  [prevappsim:med]  [prevwiz:high]  [special-delimiter-3] length tail \\
\end{tabular}
\caption{Example of plan edit over 9 edit time steps from WoW test set.  Blue are parts of the plan that were updated from the previous edit, cyan are parts that were added from the previous edit, and red are parts that get later deleted in the next edit. }
\label{tab:planeditexamples}
\end{table*}

\section{Experimental Training Details}
\label{sec:expts-training-appendix}

\subsection{Noisy Plans} Our initial experiments showed that the PLEDGE model learns to over-rely on some of the generated plan attributes, ignoring the provided dialogue history and evidence. This especially hurts the response quality in cases when the generated content plans are insufficient or contain noise. To mitigate the common errors caused by the model, we introduce two types of noise to the ground-truth plans during training time as extra regularization. First, we \textit{drop out} attributes from the planning sequence with a probability of $p_{drop}$. Second, we \textit{randomly shuffle} the entire sequence with $p_{shuf}$ probability.

\subsection{MASKER Post-processing} We observed some tokenization and repetition errors in the content plans generated by $E_Q$, potentially due to MASKER being a non-autoregressive approach. For our case, we resort to two post-processing steps to handle these errors. For tokenization errors, we simply remove the words that are not found in the training data along with the provided conversation history and evidence, which essentially covers all ill-formed words. For repetition, we simply remove the redundant words introduced after the editing stage. % Nevertheless, future work should explore relying on autoregressive text editing approaches (CITE).

\section{Examples of Generated Responses}
\label{sec:example-model-outputs-appendix}

We provide qualitative examples of dialogue model output in Table ~\ref{tab:outputexamples}. One observation is that different models' responses are generally similar, aside from a few specific phrasing details. The differences between outputs are often not a huge edit distance from each other, and this may affect the human scores, which do not differ by a significantly large margin. One explanation could be that WoW features relatively short outputs ($\sim$1-2 sentences) and grounding evidence ($\sim$1 sentence), so models trained on this data may generate relatively similar outputs with small variations. Future work developing evaluations with finer granularity may help highlight the more nuanced differences in phrasing.

% \section{Model Output Examples}
% See Table~\ref{tab:outputexamples}
\begin{table*}[ht]
\footnotesize
    \centering
    \begin{tabular}{p{3cm}p{2cm}cp{6cm}}

conv history & evidence & model & response \\
\hline
 \multirow{10}{3cm}{[Wizard] orphans often receive care from a non-relative. were you close with your parents? 
 \newline... \newline [Wizard] she seems like a good parent - the intricacies of raising a person is a very fine art. [Apprentice] parenting is the hardest job in the world and you don't get paid or rewarded enough for it}& 
 \multirow{7}{2cm}{parenting refers to the intricacies of raising a child aside from the biological relationship.}& dodeca &
 that's true. it's not easy to raise a child aside from the biological relationship.\\
&& t5 & i agree. it's a tough job, but it's worth it. parenting refers to the intricacies of raising a child aside from the biological relationship. \\
&& ctrl& yes, there are so many intricacies of raising a child aside from the biological relationship.\\
& &\pledgekwmodel{0} &i agree, it's the intricacies of raising a child aside from the biological relationship. \\
%&& \pledgestructmodel{0} &i agree. it's a lot of work, but it's a very rewarding job. \\
&& \pledgefullmodel{0} &i agree. it's a very hard job, but i think it's worth it. \\
&& \pledgekwmodel{9} &yes, it's the intricacies of raising a person aside from the biological relationship.\\\\\\
%&& \pledgestructmodel{9} &yes, parenting refers to the intricacies of raising a child aside from the biological relationship. \\
%&& \pledgefullmodel{9} &i don't think my father gets paid enough for it. he is a good father.\\
\midrule
\multirow{7}{3cm}{[Wizard] i love classical music! it's so relaxing. it's rooted in the traditions of western culture. [Apprentice] where did classical music originate?}
& \multirow{7}{2cm}{the major time divisions of western art music are as follows:  european art music is largely distinguished from many other non-european classical and some popular musical forms by its system of staff notation, in use since about the 11th century.} 
& dodeca&the major time divisions of western art music are as follows: european art music is largely distinguished from many other non-european classical and some popular musical forms by its system of staff notation, in use since about the 11th century.\\ 
& &t5& the major time divisions of western art music are as follows: european art music is largely distinguished from many other non-european classical and some popular musical forms by its system of staff notation, in use since about the 11th century.\\
&&ctrl&european art music dates back to about the 11th century. \\
& &\pledgekwmodel{0}&i'm not sure, but european art music has been around since the 11th century. \\
%& &\pledgestructmodel{0}&the major time divisions of western art music are as follows: european art music is largely distinguished from many other non-european classical and some popular musical forms by its system of staff notation, in use since about the 11th century. \\
& &\pledgefullmodel{0}&european art music has been around since the 11th century. \\
& &\pledgekwmodel{9}&the major time divisions of western art music are as follows: european art music is largely distinguished from many other non-european classical and some popular musical forms by its system of staff notation, in use since about the 11th century.\\
%& &&the major time divisions of western art music are as follows: european art music is largely distinguished from many other non-european classical and some popular musical forms by its system of staff notation, in use since about the 11th century.\\
%& &&european art music has been around since the 11th century.\\
    \end{tabular}
    \caption{Model Output Examples on WoW test set}
    \label{tab:outputexamples}
\end{table*}

\section{Human Evaluation Annotation Format}
\label{sec:human-eval-annotation-appendix}

The main focus of our evaluation was specificity and attribution, though we included sensibleness and interestingness as complementary measures.

We ask humans to rate each example for four qualities (sensibleness, specificity, interestingness, and attribution) using definitions from Lamda \cite{lamda} and by \citet{rashkin-etal-2023-ais}.  However, there were a few points where we had to clarify or expand upon how we defined attribution and specificity.

For specificity, we were careful to instruct annotators that responses need to be more than just topically specific to the conversation but also needed to capture discourse and relevance with the previous conversation utterances. This means that the response needs to be consistent with the established conversation and follow a coherent flow from the previous utterance.  While this is implied in the original definition of specificity used by Lamda (which was that {\it this response is specific to this conversational context}), we made this a more explicit requirement in our case.

For attribution, we asked annotators to only rate the attribution for the portions of the output that were pertaining to the external world.  This is a looser requirement than the original attribution paper, which evaluated all parts of the response for attribution.  This relaxation makes allowances for generic or persona comments made by the model, like ``I don't know'' and ``I want to see that movie'', that are not meant to impart external information.  We also added a rating option for annotators to declare that an example didn't have any external information that required attribution.

\subsection{Evaluation Questions}
This is the exact phrasing for the human evaluation questions.  See Section~\ref{sec:appendix:definitions} for exact definitions of evaluation dimensions provided to annotators.

{\bf 1. Evaluate Sensibleness of the Final System Response.  (on scale of 5) }\\
Does the response make sense in the context of the conversation  \\
- Yes, it makes sense. All of the information is clear and understandable.\\
- Mostly makes sense\\
- Somewhat\\
- Mostly doesn’t make sense\\
- No, the response does not make sense. The response is unclear and/or difficult to understand.

{\bf 2. Evaluate Specificity of the Final System Response. (on scale of 5)}\\
Is the response specific to the previous conversation? \\
- Yes, it is specific.  The system response addresses the user and is appropriate to the context.\\
- Mostly specific and relevant\\
- Somewhat\\
- Mostly not specific\\
- No, the response is not specific. The response ignores the user, is redundant, generic and/or vague.\\

{\bf 3. Evaluate Interestingness of the Final System Response. (on scale of 5)}\\
Is the response interesting? \\
- Yes, it is interesting.  The system response will catch the user’s attention or arouse their interest. \\
- Mostly interesting \\
- Somewhat \\
- Mostly not interesting \\
- No, the response is not interesting. The response is dry, monotonous, or disengages the user.

{\bf 4. Evaluate Attribution of the Final System Response. (multiple-choice)}
{\it \small Note: only evaluate attribution for the parts of the system response that are sharing objective information about the world.  You do not need to check attribution for stated opinions or subjective information}\\
Is all of the objective information provided by the system response fully attributable to the source document? \\
- Yes, fully attributable.  All the factual information in the system response is supported by the document.\\
- No, not fully attributable. It includes objective-seeming  information that isn’t fully supported by the document.\\
- Not applicable. This response doesn’t share any objective information

\subsection{Definitions provided to annotators for human evaluation}
\label{sec:appendix:definitions}
\begin{itemize}
    \item Specificity: Ask yourself whether the system seems to be taking the previous conversation into account or if it seems to be ignoring the previous conversation by simply writing something vague or off-topic. A response is "specific" if it stays on-topic, is attentive to what the user has said, and avoids being vague or generic. The response is ``not specific'' if it is vague, generic, or repeats information from a prior turn. It also should be marked as ``not specific'' if it seems to be ignoring the user (abruptly changing topic; ignoring their question; etc.)
    \item Attribution: Is all of the information in this response fully attributable to the information in the document? Ask yourself: “According to this document, is this response true?” A response is fully attributable to the document if ALL of the information contained in the response can be directly supported by the document. The response does not need to be stated verbatim in the document as long as all of the pertinent information is supported in the document. If any part of the response is not attributable to information provided by the document, then select “not fully attributable”. Note: if a response contains information that is factually correct but not supported by the document, you should still mark it “not fully attributable”.
    \item Sensibleness: Is the response completely reasonable and understandable?  It’s fine if it isn’t perfectly grammatically correct as long as it would be easily understood by a human user. The response ``makes sense'' if it is cohesive and understandable. If anything seems off -- not fluent, confusing, illogical, unclear pronouns, etc. -- then rate it as Does not make sense.
    \item Interestingness: A response is "interesting" if it is likely to “catch someone’s attention” or “arouse their curiosity”. The response is “not interesting” if it is dull, not engaging, or is restating obvious information.

\end{itemize}

\section{BLEU Scores}
\label{sec:bleu-appendix}
\begin{table}[th!]
\centering
\scalebox{0.65}{
\begin{tabular}{l|cccc}
\toprule
\textbf{Model} &\multicolumn{4}{c}{\textbf{BLEU}}  \\ 
& \textbf{B1} & \textbf{B2} & \textbf{B3} & \textbf{B4} \\\midrule

Attribution-Oracle & $35.79$ & $20.06$ & $14.36$ & $11.17$ \\
Specificity-Oracle & \textcolor{red}{$\mathbf{4.96}$} & \textcolor{red}{$\mathbf{0.57}$} & \textcolor{red}{$\mathbf{0.08}$} & \textcolor{red}{$\mathbf{0.02}$} \\ \midrule \midrule
\makecell[l]{E2E (\citetalias{dinan2018wizard})} & $23.02$ & $7.37$ & $3.48$ & $1.94$ \\
\makecell[l]{Dodeca (\citetalias{shuster2020dialogue})} & \textcolor{blue}{$\mathbf{37.76}$} & \textcolor{blue}{$\mathbf{20.67}$} & \textcolor{blue}{$\mathbf{14.53}$} & \textcolor{blue}{$\mathbf{11.18}$} \\
T5 (\citetalias{raffel2020exploring}) & $35.52$ & $19.44$ & $13.71$ & $10.59$ \\
ControlCodes (\citetalias{rashkin-etal-2021-increasing}) & $33.30$ & $18.71$ & $13.46$ & $10.54$ \\ \midrule
{\textbf{Plans without Editing}} \\
\pledgekwmodel{0} & $35.10$ & $19.03$ & $13.31$ & $10.28$ \\
\pledgestructmodel{0}& $36.66$ & $19.60$ & $13.72$ & $10.51$ \\
\pledgefullmodel{0}& $33.52$ & $17.72$ & $12.25$ & $9.29$ \\ \midrule
{\textbf{Plans with Editing}} \\
\pledgekwmodel{9}& $34.03$ & $18.68$ & $13.16$ & $10.19$ \\
\pledgestructmodel{9}& $33.15$ & $18.55$ & $13.26$ & $10.32$ \\
\pledgefullmodel{9}& $30.92$ & $16.60$ & $11.67$ & $8.95$ \\ \bottomrule
\end{tabular}}
\caption{\label{tab:bleu} BLEU scores (BLEU-1, BLEU-2, BLEU-3, BLEU-4) on the seen portions of WoW test set. The worst and the best scores for each column are in \textcolor{red}{\textbf{red}} and \textcolor{blue}{\textbf{blue}} respectively.}
\end{table}
In Table~\ref{tab:bleu}, we report the BLEU scores (1 through 4) on the model outputs, with the Dodeca and Attribution-Oracle models scoring the highest in most cases. However, we note that this metric has certain limitations, being a word-overlap-based metric. We observe that the BLEU score can be amplified on this dataset by always outputting the input evidence verbatim, as in the trivial Attribution-Oracle baseline.

\section{Other Metrics: Sensibility and Interestingness}
\label{sec:extra:metrics-appendix}

\begin{table}[th!]
\centering
\begin{tabular}{l|cc}
\toprule
\multicolumn{1}{c}{\textbf{Model}} & Sensible & Interesting \\
\midrule
Dodeca & $0.846$ $\pm$ $.013$ & \textcolor{blue}{$\mathbf{0.738}$} $\pm$ $.015$ \\
T5 & \textcolor{red}{$\mathbf{0.842}$} $\pm$ $.013$ & \textcolor{red}{$\mathbf{0.697}$} $\pm$ $.016$ \\
ControlCodes & $0.844$ $\pm$ $.012$ & $0.717$ $\pm$ $.016$\\
PLEDGE-KW & \textcolor{blue}{$\mathbf{0.853}$} $\pm$ $.012$ & $0.706$ $\pm$ $.016$ \\ 
\bottomrule
\end{tabular}
\caption{\label{tab:SI} Human judgements on the seen portions of WoW test set. The worst and the best scores for each column are in \textcolor{red}{\textbf{red}} and \textcolor{blue}{\textbf{blue}} respectively.}
\end{table}
There are also many other dimensions of response quality that may be complementary to the specificity and attribution. In our human evaluations of the proposed dialogue systems, we also include measurements for sensibility and interestingness (also proposed by \citet{lamda}) though we do not focus on them as the main trade-offs discussed in this paper. Some prior work has made efforts in this space; for example, \citet{aksitov2023characterizing} quantified the trade-off between attribution and fluency, which they equated to sensibleness.

In our human evaluations, we also asked humans to evaluate sensibleness and interestingness, as a way of further exploring the ongoing challenges in dialogue evaluation.  Specifically, we ask annotators to rate the sensibility of the response (Is the semantic meaning of the response understandable?) and the interestingness (Is this response likely to be engaging or appeal to the conversation partner?) on a scale of 5.  As we see in Table~\ref{tab:SI}, these scores follow slightly different trends from the other metrics.  Sensibleness generally was scored very highly on all model types, as would be expected using most commonly used language models.  The interestingness scores of all models were generally lower than their other subscores.  %But, since the annotators are different people from the original conversation partners, it may be actually pretty difficult for them to judge whether the response would be interesting to the conversation partner.
\end{document}